\documentclass[11pt]{article}

\usepackage[margin=1in]{geometry}
\usepackage{amsmath}
\usepackage{amssymb}
\usepackage{mathtools}
\usepackage{graphicx}
\usepackage{caption}
\usepackage{subcaption}
\usepackage{url}
\usepackage{pgfplots}
\pgfplotsset{width=90mm, compat=1.9}

\usepackage[dvipsnames]{xcolor}
\usepackage{tkz-graph}

\usepackage{algpseudocode}
\usepackage{algorithm}
\algrenewcommand\algorithmicrequire{\textbf{Input:}}
\algrenewcommand\algorithmicensure{\textbf{Output:}}

\usepackage[colorlinks=true, linkcolor=blue, citecolor=blue, urlcolor=blue]{hyperref}

\begin{document}

\title{Virtual neural networks: hundreds of souls in a body}

\author{
Petr Hurtik\thanks{Corresponding author. Email: \texttt{petr.hurtik@osu.cz}}, Marek Vajgl, Zahra Alijani, Vojtech Molek \\[4pt]
\normalsize Centre of Excellence IT4Innovations, Institute for Research and Applications of Fuzzy Modeling,\\
\normalsize University of Ostrava, 30. dubna 22, Ostrava, Czech Republic
}

\date{}

\maketitle

\begin{abstract}
A new concept, termed virtual neural networks, is introduced, where the count of trainable parameters is kept constant, and scalability is attained purely through computational resources. This concept is an abstract framework that can be realized using any standard convolutional neural network. It merges siamese neural networks with a deep ensemble technique by generating numerous virtual models that share weights derived from a small set of physical models. The ensemble comprises up to hundreds of trained models simultaneously. All virtual networks take the same input, and their interconnected structure induces an internal distortion that boosts the entire ensemble robustness. The accuracy of the ensemble improves as the number of virtual networks increases, without changing the capacity. Virtual neural networks outperform larger capacity models, typical deep ensembles, and contemporary approaches like SWA and Masksembles. Additionally, the highest-performing individual model from the ensemble surpasses other models trained individually, even those with a greater number of parameters.
Code: gitlab.com/EnginCZ/virtual-models-public

\end{abstract}

\begin{figure}
\centering
\includegraphics[width=120mm]{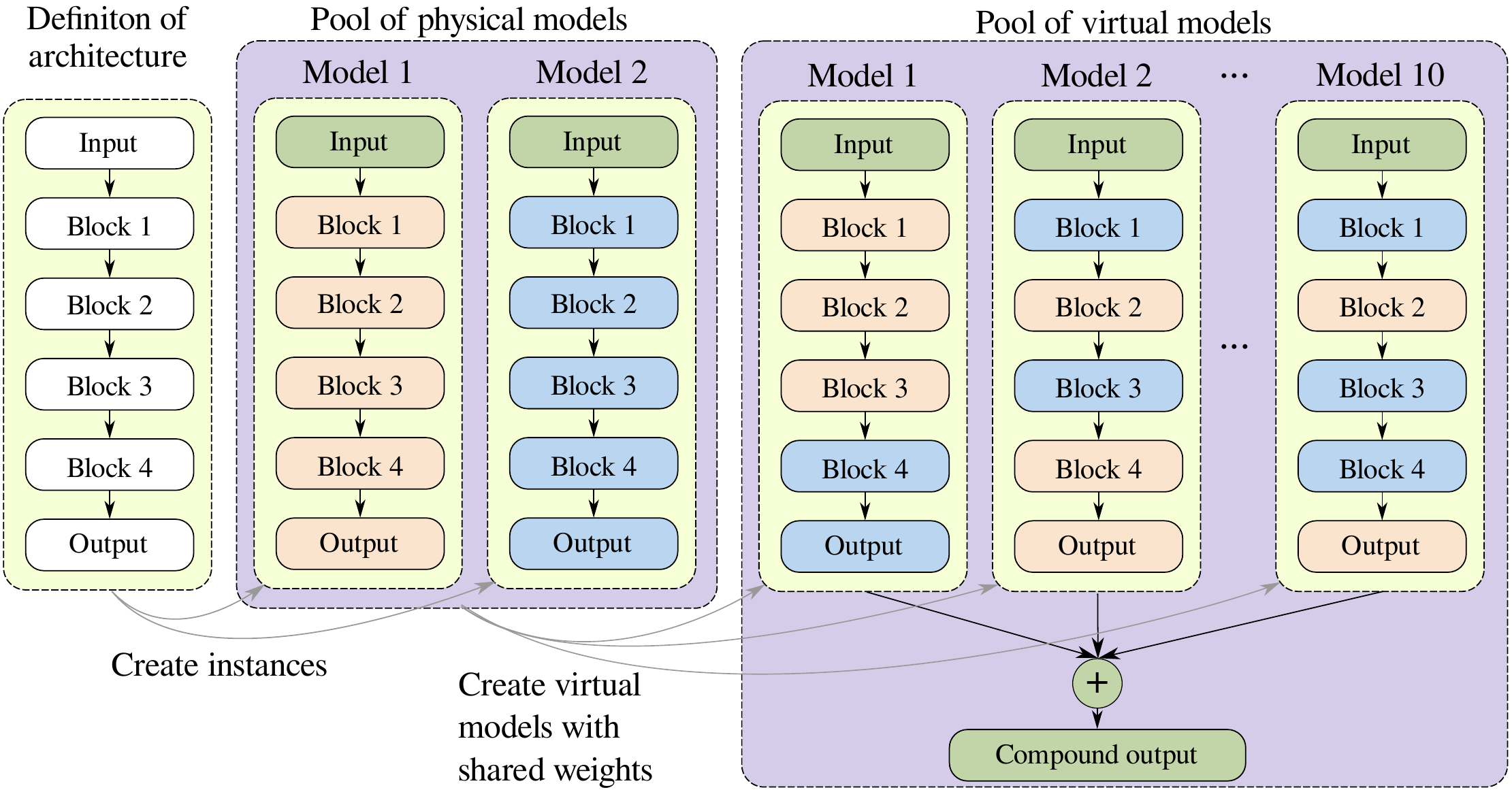}
\caption{Diagram of the proposed scheme for two physical and ten virtual models. The leftmost column defines the prototypical model architecture. The architecture is used to instantiate models in a physical model pool. Each virtual model shares its blocks (weights) with random physical model blocks at the same level, creating unique combinations.}
\label{fig:main_scheme}
\end{figure}

The principle of assembling diversified weaker classifiers into a stronger classifier is a basic idea in machine learning used in Boosting~\cite{breiman1998arcing} technique and also reflected in Stacking~\cite{wolpert1992stacked}, Bagging~\cite{breiman1996bagging}, or Bayesian model average~\cite{hoeting1999bayesian} / combination~\cite{monteith2011turning}.  Simple classifiers can also be replaced by neural networks to create a deep ensemble~\cite{lakshminarayanan2016simple} to boost their performance even further.
The advantages of an ensemble are mainly reduced overfitting~\cite{hashem1997optimal} and lower complexity compared to a single model with the same precision~\cite{pearlmutter1990chaitin}. It has been shown~\cite{buschjager2020generalized} that when an error is considered as a composition of variance and bias, the ensemble reduces variance while keeping bias low. The real usefulness of a simple averaging of predictions obtained by several models was also proven in the AlexNet paper~\cite{krizhevsky2012imagenet} and is currently the must for winning solutions on the Kaggle competition platform. 

The practical problem arises when the ensemble consists of too many models, such as 37 models in the winning solution of the Carvana Masking Challenge~\cite{Carvana2017}. The models are too large and cannot be placed into GPU memory at once, so to make inference for a single input, the models have to be loaded/deleted sequentially, which significantly increases the inference time. In addition, the models are trained separately (due to the fact that they cannot be placed into GPU VRAM at once), so their diversity is not guaranteed. It means that the outputs of the partial networks can be correlated, and thus, their ensemble does not lead to an increase in accuracy; for more details see~\cite{liu2019deep}. The only way to control diversity here is to involve different data or architectures. The aim of our work is to address these disadvantages.\emph{Virtual Neural Networks Ensemble (VNNE)} is proposed as an implicit homogeneous ensemble that combines the principles of siamese neural networks~\cite{chicco2021siamese} with the deep ensemble technique~\cite{lakshminarayanan2016simple}. \textbf{Thanks to the virtual shared parameters, VNNE enables the concurrent training and inference of hundreds of neural networks without the need to load or delete them sequentially from GPU memory.} This characteristic inspires the name "hundreds of souls in a body." Additionally, experimental results demonstrate that, with the same number of trainable parameters, VNNE outperforms standard deep ensembles.

\noindent\textbf{The highlights of the paper are as follows:}
\begin{itemize}
    \item A novel approach to assembling is introduced, where a large set of tangled virtual neural networks share weights based on the siamese neural network principle, utilizing a few physical neural networks following the deep ensemble principle; see Figure~\ref{fig:main_scheme}.
    \item Unlike standard neural networks, the virtual capacity is scaled while the number of trainable parameters remains the same, as shown in Figure~\ref{graph-ablation-c}.
    \item The rationale behind VNNE is that the tangled structure of virtual networks forms an inner distortion; read more in Section~\ref{subsec-rationale}.
    \item Models in VNNE have lower disagreement diversity and higher Kullback-Leibler divergence than models in the standard deep ensemble, i.e., they are less overconfident in wrong predictions, as VNNE have higher accuracy; see Table~\ref{table-cifar-resnet}.
    \item The best single model taken from VNNE achieves better results compared to other single-trained models of the same architecture. The same holds even compared to models of architecture with higher capacity.
\end{itemize}

\section{Related work}
\textbf{Ensembling.} Training and testing costs are one of the most important constraints for ensembles. To address this issue, various approaches have been proposed.
LegoNet~\cite{yang2019legonet} uses lightweight filters that are combined into standard-size filters that share the parameters. FGE \cite{garipov2018loss} uses the fact that the surface of a complex loss function is connected by curves over which training and test accuracy are nearly constant and proposes an implicit set of models on the same path between nodes. To avoid the FGE's issue of increasing the computation time of inference due to making inference of $n$ models, SWA~\cite{izmailov2018averaging} stores model's weights during the $n$ last epoch and averages them into a single model. Wand et al.~\cite{wang2021memory} showed that such averaging can be used efficiently with weight pruning. The implicit ensemble of SWA is changed to explicit in SWAP~\cite{gupta2020stochastic} by training several models in parallel and averaging their weights after several epochs repeatedly. A similar idea of the implicit ensemble is proposed in~\cite{huang2017snapshot} where a model is trained using the cyclic learning rate~\cite{loshchilov2016sgdr}, its weights (snapshots) are stored before the learning rate is restarted, and then the stored weights are averaged. Compared to SWA, the Snapshot ensemble weights are more diverse. Diversity refers to the situation in which individual models within an ensemble disagree when they are incorrect and do not consistently predict the same class. Given two ensembles with the same accuracy, the one with greater model diversity is more useful, as this diversity can be leveraged to measure the uncertainty of the ensemble.
The heterogeneous explicit ensemble is used in \cite{zhang2020diversified}, where the outputs of several models are passed through shared layers to produce the final output (module) for each model. The modules are then assembled. Takahashi et al.~\cite{takahashi2018novel} takes a base network and creates two copies of it. The copies share weights and are continuously downsampled by factor 2 to the input resolution. The aggregated outputs are used to make the final prediction. Havasi et al. \cite{havasi2020training} use the MIMO approach, in which by feeding a network with $n$ various images and taking $n$ outputs, the network must create subnets internally. During testing, only one image is used, copied $n$ times, and the output is averaged. They found that the ideal number of subnetworks is about 2-3 only.

A specific loss function for the ensemble is tackled by negative correlation learning~\cite{brown2005managing}, generalized and negative correlation learning~\cite{buschjager2020generalized}, evaluation of cross-entropy between the ensembled models~\cite{opitz2016efficient}, the addition of Kullback-Leibler diversity~\cite{webb2019joint}, or by incorporating both diversity and cooperation losses at the same time~\cite{dvornik2019diversity}. The issue of the described ensemble is that all models in the ensemble are trained jointly; it leads to increase of of memory requirements and does not allow train big ensembles.

The entire research direction is devoted to dropout \cite{srivastava2014dropout,baldi2013understanding} as a form of the ensemble technique and a representative of pseudo-ensemble class~\cite{bachman2014learning}. The class is characterized by a parent model, from which the child models are sampled with respect to a certain noise type.  Monte Carlo dropout, as employed by Gal et al.~\cite{gal2016dropout}, allows dropout to be applied during inference. To control the process of generating subnetworks, Durasov et al. \cite{durasov2021masksembles} create random dropout subnetworks by defining fixed dropout masks at the beginning of training and making predictions using the models given by the masks. Kobayashi et al. \cite{kobayashi2022diverse} and Liu et al. \cite{liu2021deep} call the models created by a mask as ticket and show that from a pre-trained model, a diversified ensemble can be established by involving distinct tickets.  Masks can also be created in a guided scheme to obtain non-overlapping subnetworks~\cite{zhang2021ex}. The disadvantages of this research direction are the decreased capacity of the base network and the increased computation time during inference. The crucial decision is how to select the appropriate dropout rate: the subnetworks are too similar, and the diversity is low when the dropout rate is low. When the dropout rate is higher, capacity is reduced, particular subnetworks perform worse, and thus the performance of the ensemble is also lower.

\textbf{Siamese neural network}. The original idea~\cite{bromley1993signature,chicco2021siamese} is based on several identical neural networks that fully share weights and are trained to produce similar latent representations for similar inputs and vice versa, i.e., its aim is semantic similarity. Learning is carried out using a simple Euclidean / cosine distance or contrastive loss~\cite{wang2021understanding}. In the area of image processing, applications include the recognition of signatures~\cite{bromley1993signature}, faces~\cite{taigman2014deepface}, or gestures~\cite{yang2019learning}, to name a few. Some publications~\cite{ferreira2021ensembling,huang2022ensemble} focus on the assembly of siamese neural networks, but are restricted to the standard task of matching latent representations. BatchEnsemble~\cite{wen2020batchensemble} shares weights between all members of the ensemble, which are taken and multiplied by vectors whose weights are private for a certain member. Such a scheme saves the required memory due to shared main weights.

The proposed technique, VNNE, combines the siamese neural network approach with a deep ensemble. A novel partial siamese neural network is introduced, consisting of virtual models that are mutually tangled and share the weight space of the physical models; see Figure~\ref{fig:main_scheme}. Unlike a typical siamese network, the network output is not a latent representation but class probabilities collectively aggregated by a non-trainable aggregation layer in the final ensemble prediction. The aggregation layer, which lacks trainable parameters, enables the selection of a single virtual network from a trained VNNE for use as a standalone standard model.

\textbf{VNNE vs ensembles}. VVNE is not a traditional ensemble method.  
\textbf{1)} Subnetworks are not created by reducing the parameters of a single base network (e.g., through dropout or masks); instead, all subnetworks maintain the same number of parameters as the base network.  
\textbf{2)} The overall ensemble size grows while the memory required for training increases sublinearly.  
\textbf{3)} All subnetworks are optimized jointly.  
\textbf{4)} At any stage of optimization, all subnetworks are ready for use.

\section{Virtual neural networks ensemble (VNNE)}
In an ensemble of models, the key principle in achieving good performance is the diversity between the involved models \cite{liu2019deep,wu2021boosting}, where the diversity itself can be an optimization criterion during the training procedure \cite{zhang2021ex}. Diversity refers to a low correlation between the produced probabilities. While an ensemble of models enhances generalization, it also results in increased size and reduced computational speed. In various competitions on platforms such as Kaggle and Signate, ensembles often consist of more than ten models, making them unsuitable for deployment in end-user applications.
Motivated by the work described in \cite{durasov2021masksembles,zhang2021ex,havasi2020training}, where a single base network includes subnetworks, this study explores a scheme that incorporates multiple base networks to combine the advantages of traditional deep ensembles~\cite{lakshminarayanan2016simple} and the theory of subnetworks.

In a box, the main principle of VNNE is to build subnetworks from the parts of the base networks. First, a basic architecture (e.g., ResNet) is chosen and split into a logical sequence of disjoint blocks where a block is represented by a single or multiple layers. Then, several standard neural networks (called \textit{physical}) are created as usual with respect to the architecture. Finally, the derived networks (called \textit{virtual}) are constructed as follows. For every block of every virtual network in the architecture, the corresponding block is taken from the randomly selected physical network (see Figure \ref{fig:main_scheme}). The final VNNE is loaded into the GPU at once and all virtual networks are trained jointly. This means that the diversity of models is a natural product of the training procedure if needed. In other words, the process of creation put the weights of the physical layers into the memory. So, when virtual neural networks are created, the layers with their weights already exist in memory, and the composition of VNNES establishes logical connections between these weights. When multiple VNNE share the same weight, this weight is not duplicated. The following section describes the whole construction procedure in more detail.

\subsection{Construction of virtual neural networks ensemble}
\textbf{A physical neural network} used for weight sharing in virtual networks can be formalized as a mapping of a three-dimensional tensor to a one-dimensional vector. For simplicity, we omit batch dimension and consider the task of classifying images into $\ell$ classes, but VNNE can also be adapted for image segmentation, image detection, image generation, etc.
Consider a pool $\mathcal{P}=\{P_1, P_2, \dots, P_p\}$ composed of $p\geq2$ \emph{physical} neural networks. For simplicity, let $p=2$. Each network operates as a function $P_i: \mathbb{R}^{h\times w \times 3} \to \mathbb{R}^\ell$, where a standard raster color image of resolution $h\times w$ is considered as input.
 Moreover, each of the physical networks consists of $b > 0$ \emph{blocks}, that is, $P_i = \{B_{i,1}, B_{i,2}, \dots, B_{i,b}\}$. Here, we suppose that a block is an abstract structure whose internal composition is given by a particular neural network architecture. 
In the case of CNN, a block can consist of a sequence of convolutional, batch normalization, and activation layers. Each block, except $B_{i,b}$, realizes $B_{i,j}: \mathbb{R}^{x \times y \times z} \to \mathbb{R}^{x' \times y' \times z'}$ where the exact value of the dimensions depends on the definition of layers (number and size of filters) within the block. The last block realizes $B_{i,b}: \mathbb{R}^{x \times y \times z} \to \mathbb{R}^\ell$. The necessary condition for the described ensemble is that $x, y, z$, and $x', y', z'$ are equal for all blocks at the same level $j$. This is naturally satisfied by the fact that VNNE is a homogenous ensemble because all instances of physical networks are constructed according to the same origin architecture; see Fig.~\ref{fig:main_scheme}. The fact also leads to the following: for arbitrary $i, i'$ and $j$  where $i$ and $i'$ are the indices of the physical neural network and $j$ is the block index, holds $c(B_{i, j}) = c(B_{i', j})$, where $c$ stands for capacity, i.e., the number of unique trainable parameters. From this, $c(P_i) = c(P_{i'})$ is also valid. Physical neural networks are not used for training or inference and serve only as placeholders for weights that are shared between virtual networks. 

\textbf{Virtual neural networks.} Formally, a pool of $v \geq p$ virtual neural networks is defined as $\mathcal{V} = \{V_1, V_2, \dots, V_v\}$.  
Similar to physical networks, each virtual network consists of $b$ blocks. A virtual network is then defined as  
$V_i = \{B_{r(1,p),1}, B_{r(1,p),2}, \dots, B_{r(1,p),b}\}$, where $r(1, p)$ is a random integer sampled from a discrete uniform distribution over $\{1, \dots, p\}$.  

The term \emph{virtual} is used because each $V_i$ is composed of the weights of the physical networks and does not contain its own trainable parameters. In other words, the condition $c(\mathcal{P}) \geq c(\mathcal{V})$ holds, meaning that increasing $v$ does not lead to an increase in the number of trainable parameters. Due to the random selection of physical blocks used to construct virtual networks, there may be instances where one or more physical blocks are not included in the ensemble. Consequently, the final capacity satisfies $c(\mathcal{V}) \in [c(P_i), c(\mathcal{P})]$, where $P_i$ is an arbitrary network from the pool.  

However, in all experiments conducted, $c(\mathcal{V}) = c(\mathcal{P})$ was consistently observed; see Table~\ref{table-cifar-resnet} and Table~\ref{table-cifar-effnet}.

Finally, let $f: \mathbb{R}^{v \times \ell} \to \mathbb{R}^\ell$ denote the output layer of the proposed homogeneous implicit ensemble. In practice, $f$ can be an arbitrary aggregation function such as mean, max, or product. As $f$, we consider the averaging $f_A(\cdot) = 1/v\sum_{i=1}^{v}V_i(\mathbf{X})$ and the multiplication (product) $f_M(\cdot)= \prod_{i=1}^{v} V_i(\mathbf{X})$, where $\mathbf{X}$ is the set of input data. The disadvantage of multiplication
is $0 \approx \prod_{i=1}^{v} V_i(\mathbf{X})$ when $\exists i: V_i(\mathbf{X}) \approx 0$.
This makes the training process unstable because the output of one virtual network can nullify the predictions of all other virtual networks for a particular class. 
However, the effect of multiplication emphasizes the virtual subnetwork with the highest loss value.

The described algorithm allows the construction of unique $p^b$ virtual neural networks. For a step-by-step construction, refer to Algorithm~\ref{alg:vnc}. Unlike other publications proposing weight sharing \cite{takahashi2018novel,yang2018unsupervised,zhang2020diversified}, virtual networks do not have private layers with non-shared parameters, and there is no trainable aggregation layer.

\begin{algorithm}[!ht]
\small
\caption{How to create the structure of the virtual neural network ensemble.}
\label{alg:vnc}
\begin{algorithmic}[1]
\Require 
\Statex Number of physical networks $p \geq 2$
\Statex Number of virtual networks $v \geq p$
\Statex Number of blocks based on the underlying network architecture $b > 0$
\Statex Aggregating function $f$
\Ensure
\Statex List of physical networks $\mathcal{P}$
\Statex List of virtual networks $\mathcal{V}$
\Statex Model of VNNE for training $\mathcal{M}$\medskip

\For{$i = 1, ..., p$}
    \For{$j = 1, ...,  b$}
        \State $B_j \gets$  create\_network\_block(j) 
        \Comment{build blocks for physical network}
    \EndFor
    \State $\mathcal{P}_i \gets$ compile\_model($B$)
    \Comment{assemble blocks into a physical network model}
\EndFor

\For{$i = 1, ..., v$}
    \For {$j = 1, ..., b$}
        \State $k \gets$ random(1, $p$)
        \State $L_j \gets \mathcal{P}_{k,j}$ 
        \Comment{get $j^{th}$ block of a random $k^{th}$ physical model}
    \EndFor
    \State $\mathcal{V}_i \gets$ compile\_model($L$)
    \Comment{assemble blocks into a virtual network model}
\EndFor
\State $\mathcal{M} \gets$ aggregate\_models($\mathcal{V}, f$)
\end{algorithmic}
\end{algorithm}

\subsection{Loss function}
Given a set of input data $\mathbf{X}$ and the corresponding labels $\mathbf{Y}$, we define a base loss function $\mathcal{L}_B$ as 
\begin{equation}
\mathcal{L}_B(\mathbf{X}, \mathbf{Y}) = \mathcal{H}(f(V_1(\mathbf{X}), V_2(\mathbf{X}), \dots, V_v(\mathbf{X})), \mathbf{Y}),
\end{equation}
where $\mathcal{H}$ represents an arbitrary loss function such as cross-entropy or focal loss for image classification, or Dice loss in the case of image segmentation, and $f$ represents an aggregation function. Based on our observation, training with the mentioned $\mathcal{L}_B$ converges very slowly, and the overall performance is low; Training many virtual networks (we tested mainly ten and twenty virtual networks) simultaneously seems too complicated. Therefore, we propose to strengthen the training by involving the loss of particular virtual networks and modifying $\mathcal{L}$ in the form of
\begin{equation}
\mathcal{L}(\mathbf{X}, \mathbf{Y}) = \mathcal{L}_B(\mathbf{X}, \mathbf{Y}) + \omega \sum_{i=1}^{v}\mathcal{H}(V_i(\mathbf{X}), \mathbf{Y}),
\label{full-loss}
\end{equation}
where $\omega$ is a positive constant whose certain value is the aim of hyperparameter optimization. When $\omega$ is too large, the loss of virtual networks will suppress the loss of the ensemble, resulting in poor performance. When $\omega$ is too low, the effect will be small, resulting (again) in poor performance. See Section~\ref{sec-ablation_study} for the proper selection of $\omega$ and its impact. The described loss is similar to that in~\cite{zhang2020diversified}, with the difference that the loss of partial virtual networks is weighted, and the diversity loss is not used. The rationale for this choice is provided in Section~\ref{subsec-diversity-loss}.

\subsection{Updating shared parameters}
Let us illustrate the shared parameter update process in VNNE of two ($v=2$) virtual networks $V(x) = \frac{1}{v}V_1(x) + \frac{1}{v}V_2(x)$ using an average aggregation function. For the sake of simplicity, let the virtual networks be composed of three simple perceptrons without bias, a single scalar value input $x$, and a scalar target $y$. Their respective outputs are $V_1(x) = \sigma(\sigma(\sigma(xw_0)w_1)w_3)$ and $V_2(x) = \sigma(\sigma(\sigma(xw_0)w_2)w_3)$ where $w_i$ is the scalar weight. We will assume that the activation function $\sigma(a) = a$ is an identity, thus removing it from the equations for clarity reasons: $V_1(x) = xw_0w_1w_3$ and $V_2(x) = xw_0w_2w_3$. We will use the difference: $\mathcal{L}(x,y) = x - y$ for a loss function. $V_1$ and $V_2$ definitions show that both networks share common weights $w_0$ and $w_3$. To update shared parameters, we will use stochastic gradient descent in its simple form without learning rate $\alpha$:
\begin{equation}
    w_i = w_i - \frac{ \partial \mathcal{L}(V(x), y) }{\partial w_i}.
\end{equation} 
The following equation shows the result of partial derivation with respect to the shared weights:
\begin{equation*}
    \begin{aligned}
        \frac{ \partial \mathcal{L}(V(x), y) }{\partial w_0} & =\frac{\partial (\frac{1}{2}V_1(x) + \frac{1}{2}V_2(x) - y)}{\partial w_0} \\
       & =\frac{\partial (\frac{1}{2}(xw_0w_3)(w_1+w_2)- y)}{\partial w_0} \\
        & =\frac{1}{2}xw_3(w_1+w_2)
        \end{aligned}
        \end{equation*}
       and
       \begin{equation*}
        \frac{ \partial \mathcal{L}(V(x), y) }{\partial w_3}  = \frac{1}{2}xw_0(w_1 + w_2).
\end{equation*}
 The partial derivations show that both the virtual networks $V_1$ and $V_2$ contribute to the update. In more general terms, all virtual networks within the ensemble that contain a shared block $B$ contribute to the update of the parameters $B$. The update contribution ratio of the networks ($\frac{1}{2}$ in the example case) depends on the aggregation function. The situation where more networks contribute to update a certain shared parameter, i.e., the update is a combination of partial updates, is analogous to situation of increasing batch size where the update is realized with respect to more input data. As is known from the literature~\cite{he2019control}, increasing batch size (with appropriate learning rate) leads to increased generalization, so it is assumed that increasing the number of virtual neural networks may have a similar effect. This is numerically validated in Ablation study.

\subsection{The rationale behind}
\label{subsec-rationale}

\begin{figure}[!t]
    \centering
    \includegraphics[width=40mm]{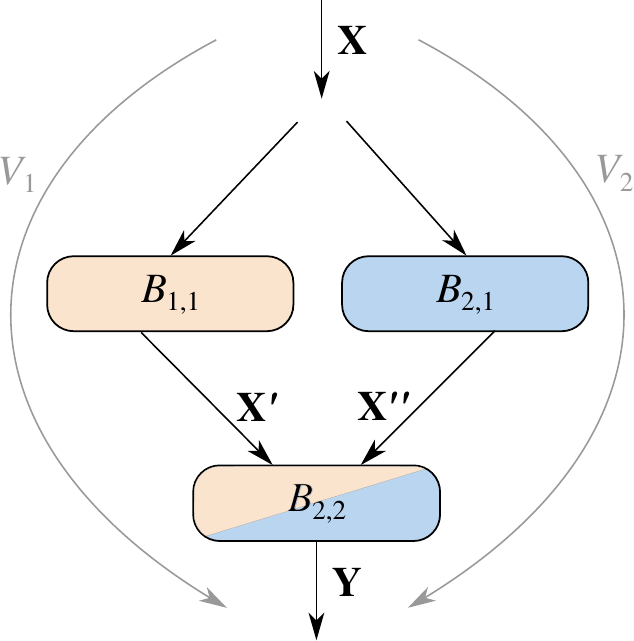}
    \caption{Two simple virtual networks $V_1 = \{B_{1,1}, B_{1,2}\}$ and $V_2 = \{B_{2,1}, B_{1,2}\}$ sharing a common block $B_{1,2}$. The networks have same input $\mathbf{X}$ and produce a similar output $B_{1,2}(\mathbf{X'}) \approx B_{1,2}(\mathbf{X''}) \approx \mathbf{Y}$.}
     \label{fig-scheme1}
\end{figure}

The necessary condition of the rationale is that we have $p, v \geq 2$. Let us suppose a simple scheme, as illustrated in Fig.~\ref{fig-scheme1}. It consists of two virtual networks $V_1 = \{B_{1,1}, B_{1,2}\}$ and $V_2 = \{B_{2,1}, B_{1,2}\}$ that are made from at least two physical neural networks and includes three blocks at two levels. The ensemble takes the input data $\mathbf{X}$ and processes them in blocks $B_{1,1}$ and $B_{2,1}$. Here, we consider, for illustration, that these two blocks are derived from distinct physical networks, and thus do not share weights. Because the weights of the blocks were randomly initialized, we can write $B_{1,1}(\mathbf{X}) = \mathbf{X'}$, $B_{2,1}(\mathbf{X}) = \mathbf{X''}$ where $\mathbf{X'} \neq \mathbf{X''}$. By training these virtual networks together, the shared block $B_{1,2}$ becomes more resistant to its input, which is a combination of $\mathbf{X', X''}$. Since these blocks must accommodate multiple possible feature transformations from different virtual networks, they are less sensitive to overfitting on specific input patterns, leading to improved robustness against noise or adversarial perturbations.

Furthermore, we consider that both virtual networks involve the block $B_{2,2}$. This means that the block refers to \emph{the same} physical weights, and therefore the weights between virtual networks are shared, identical, and updated by both virtual blocks with the same priority. Because it is the last block of our simple network and produces the probabilities (we here omit the aggregation layer for simplicity), the network is trained to produce $B_{1,2}(\mathbf{X'}) \approx B_{2,2}(\mathbf{X''}) \approx \mathbf{Y}$. In other words, the block $B_{1,2}$ is trained to accept two different sets of feature maps (produced by the same data and various preceding blocks) and to produce the same prediction. This forces the block $B_{2,2}$ to have its internal weights robust enough with respect to perturbations in its input, i.e., the inner combination. Note that the described scheme is similar to multi-path blocks in ResNeXt~\cite{xie2017aggregated} with the difference that ResNeXt does not use shared weights and utilizes the aggregation layer after each block. In our case, we share weights among random blocks, realize inner aggregation by putting multiple outputs from distinct physical blocks into one randomly, and have only one fixed aggregation layer at the end of our ensemble.

Our scheme resembles data augmentation, with the difference that the distortion here is created by different weights in the virtual blocks $B_{1,1}$ and $B_{2,1}$, which can be written as $Y \approx B_{1,2}(B_{1,1}(\mathbf{X})) \approx B_{1,2}(B_{2,1}(\mathbf{X}))$. Clearly, the deeper the network and the more virtual networks we have, the stronger the combination/distortion. We hypothesize that such inner combination of feature maps leads to the production of more robust weights to process more general feature maps, which decrease overfitting and thus improve the performance of such a model.


\subsection{Collapsing into the same network?}
\begin{figure}[!h]
    \centering
    \begin{tikzpicture}
    \begin{axis}[legend columns=4, height=40mm,width=80mm, xlabel={Epoch},ylabel={Kullback-Leibler diversity},xmin=1, xmax=100, ymin=0.0, ymax=1.0,
    ymajorgrids=true,yminorgrids=true,style={thick},legend style={at={(1.0,0.97)}},axis x line*=bottom,axis y line*=left, style={font=\footnotesize}]
    \addplot[color=green, smooth]coordinates{(1,1)(2,0.849)(3,0.761)(4,0.699)(5,0.651)(6,0.611)(7,0.577)(8,0.548)(9,0.523)(10,0.5)(11,0.479)(12,0.46)(13,0.443)(14,0.427)(15,0.412)(16,0.398)(17,0.385)(18,0.372)(19,0.361)(20,0.349)(21,0.339)(22,0.329)(23,0.319)(24,0.31)(25,0.301)(26,0.293)(27,0.284)(28,0.276)(29,0.269)(30,0.261)(31,0.254)(32,0.247)(33,0.241)(34,0.234)(35,0.228)(36,0.222)(37,0.216)(38,0.21)(39,0.204)(40,0.199)(41,0.194)(42,0.188)(43,0.183)(44,0.178)(45,0.173)(46,0.169)(47,0.164)(48,0.159)(49,0.155)(50,0.151)(51,0.146)(52,0.142)(53,0.138)(54,0.134)(55,0.13)(56,0.126)(57,0.122)(58,0.118)(59,0.115)(60,0.111)(61,0.107)(62,0.104)(63,0.1)(64,0.097)(65,0.094)(66,0.09)(67,0.087)(68,0.084)(69,0.081)(70,0.077)(71,0.074)(72,0.071)(73,0.068)(74,0.065)(75,0.062)(76,0.06)(77,0.057)(78,0.054)(79,0.051)(80,0.048)(81,0.046)(82,0.043)(83,0.04)(84,0.038)(85,0.035)(86,0.033)(87,0.03)(88,0.028)(89,0.025)(90,0.023)(91,0.02)(92,0.018)(93,0.016)(94,0.013)(95,0.011)(96,0.009)(97,0.007)(98,0.004)(99,0.002)(100,0)};\addlegendentry{S1~~~~}
    \addplot[color=blue, smooth]coordinates{(1,1)(2,0.85)(3,0.764)(4,0.718)(5,0.681)(6,0.65)(7,0.623)(8,0.6)(9,0.58)(10,0.561)(11,0.545)(12,0.529)(13,0.516)(14,0.503)(15,0.491)(16,0.48)(17,0.469)(18,0.459)(19,0.45)(20,0.441)(21,0.433)(22,0.425)(23,0.418)(24,0.41)(25,0.404)(26,0.397)(27,0.391)(28,0.385)(29,0.379)(30,0.373)(31,0.368)(32,0.362)(33,0.357)(34,0.352)(35,0.348)(36,0.343)(37,0.339)(38,0.334)(39,0.33)(40,0.326)(41,0.322)(42,0.318)(43,0.314)(44,0.311)(45,0.307)(46,0.303)(47,0.3)(48,0.297)(49,0.293)(50,0.29)(51,0.287)(52,0.284)(53,0.281)(54,0.278)(55,0.275)(56,0.272)(57,0.269)(58,0.267)(59,0.264)(70,0.25)(71,0.25)(72,0.25)(73,0.25)(74,0.25)(75,0.25)(76,0.25)(77,0.25)(78,0.25)(79,0.25)(80,0.25)(81,0.25)(82,0.25)(83,0.25)(84,0.25)(85,0.25)(86,0.25)(87,0.25)(88,0.25)(89,0.25)(90,0.25)(91,0.25)(92,0.25)(93,0.25)(94,0.25)(95,0.25)(96,0.25)(97,0.25)(98,0.25)(99,0.25)(100,0.25)
};\addlegendentry{S2~~~~}
    \addplot[color=red, smooth]coordinates{(1,1)(2,0.784)(3,0.687)(4,0.618)(5,0.565)(6,0.522)(7,0.485)(8,0.453)(9,0.425)(10,0.399)(11,0.377)(12,0.356)(13,0.337)(14,0.319)(15,0.303)(16,0.287)(17,0.273)(18,0.259)(19,0.246)(20,0.24)(21,0.235)(22,0.231)(23,0.229)(24,0.227)(25,0.225)(26,0.224)(27,0.225)(28,0.226)(29,0.227)(30,0.228)(31,0.23)(32,0.24)(33,0.255)(34,0.27)(35,0.285)(36,0.3)(37,0.32)(38,0.34)(39,0.355)(40,0.36)(41,0.365)(42,0.367)(43,0.369)(44,0.372)(45,0.375)(46,0.378)(47,0.38)(48,0.384)(49,0.385)(50,0.385)(51,0.385)(52,0.386)(53,0.386)(54,0.387)(55,0.387)(56,0.387)(57,0.388)(58,0.388)(59,0.389)(60,0.389)(61,0.389)(62,0.39)(63,0.39)(64,0.39)(65,0.391)(66,0.391)(67,0.391)(68,0.392)(69,0.392)(70,0.392)(71,0.393)(72,0.393)(73,0.393)(74,0.393)(75,0.394)(76,0.394)(77,0.394)(78,0.395)(79,0.395)(80,0.395)(81,0.395)(82,0.396)(83,0.396)(84,0.396)(85,0.396)(86,0.397)(87,0.397)(88,0.397)(89,0.397)(90,0.398)(91,0.398)(92,0.398)(93,0.398)(94,0.399)(95,0.399)(96,0.399)(97,0.399)(98,0.4)(99,0.4)(100,0.4)};\addlegendentry{S3~~~~}
    \addplot[color=black, smooth]coordinates{(1,0.669)(2,0.655)(3,0.642)(4,0.644)(5,0.643)(6,0.645)(7,0.638)(8,0.634)(9,0.633)(10,0.637)(11,0.623)(12,0.614)(13,0.599)(14,0.6)(15,0.6)(16,0.603)(17,0.608)(18,0.609)(19,0.599)(20,0.602)(21,0.603)(22,0.608)(23,0.601)(24,0.581)(25,0.574)(26,0.577)(27,0.573)(28,0.563)(29,0.546)(30,0.518)(31,0.505)(32,0.485)(33,0.484)(34,0.485)(35,0.464)(36,0.438)(37,0.415)(38,0.402)(39,0.397)(40,0.384)(41,0.364)(42,0.356)(43,0.351)(44,0.335)(45,0.331)(46,0.309)(47,0.306)(48,0.284)(49,0.267)(50,0.255)(51,0.251)(52,0.244)(53,0.238)(54,0.229)(55,0.223)(56,0.219)(57,0.209)(58,0.204)(59,0.201)(60,0.199)(61,0.194)(62,0.187)(63,0.183)(64,0.176)(65,0.174)(66,0.168)(67,0.168)(68,0.167)(69,0.165)(70,0.163)(71,0.161)(72,0.16)(73,0.158)(74,0.157)(75,0.155)(76,0.154)(77,0.153)(78,0.153)(79,0.152)(80,0.152)(81,0.15)(82,0.15)(83,0.15)(84,0.15)(85,0.15)(86,0.151)(87,0.151)(88,0.152)(89,0.151)(90,0.151)(91,0.151)(92,0.151)(93,0.151)(94,0.152)(95,0.153)(96,0.153)(97,0.153)(98,0.153)(99,0.153)(100,0.152)};\addlegendentry{Measured}
    \end{axis}
    \end{tikzpicture}
    \caption{Experiment monitoring collapsing of virtual networks into the same networks expressed by KL diversity. S1, S2, and S3 visualize three possible scenarios,  \emph{Measured} shows the real situation during training. The experiment proves that the network does not converge into a collapsed state.}
    \label{fig-graph-divergence-scenario}
\end{figure}
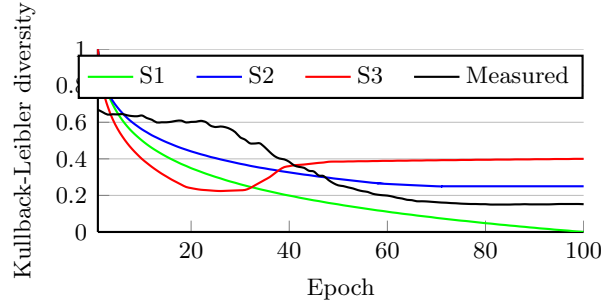

Speaking in terms of general neural networks, we mark as \emph{collapsed} such a state, where $\forall(i, i'): B^W_{i, j} = B^W_{i', j}$, for an arbitrary $j$ and where $B^W_{i, j}$ denotes weights of the block. Speaking of the ensemble, we are more interested in outputs than in weights, so we define a collapsed state as $\forall(i, i'): V_i(\cdot) = V_{i'}(\cdot)$. To express the inverse degree of collapse, the Kullback-Leibler diversity ($d_{KL}$)~\cite{liu2021deep} can be used. We suppose three possible scenarios:
\begin{enumerate}
    \item[S1]: $d_{KL}$ will decrease through training until it is equal to zero, i.e., all virtual networks will collapse into a single network.
    \item[S2]: $d_{KL}$ will decrease, but the decrease will eventually stop. A further decrease in divergence would lead to an increase in the loss of the whole ensemble, and this prevents it from collapsing.
    \item[S3]: $d_{KL}$ will first decrease, stabilize at some point, and increase until an equilibrium is reached. Here, the increase is supposed to be the only case in which the performance of the ensemble increases when increasing the performance of particular networks does not lead to further improvement in the loss.
\end{enumerate} 
These three scenarios are visualized in Figure~\ref{fig-graph-divergence-scenario}. We empirically estimated real progress by training VNNE for two physical and 20 virtual NNs with aggregation by multiplication and measuring $d_{KL}$ after each epoch of training on CIFAR-10 classification task. The results given by the graph in Figure~\ref{fig-graph-divergence-scenario} show that it converges to an equilibrium similar to S2, that is, a stable state without collapsing.

In the area of deep neural networks, there are two similar situations known that do not end with collapse. a) there are multiple trainable convolutional kernels in a layer in standard convolutional neural networks where the kernels do not collapse into the same state regardless of whether they process the same input data; b) identically structured multi-paths known from ResNeXt \cite{xie2017aggregated} that process the same input and are aggregated at their end; here, the scheme is functional as well without its collapsing. The necessary condition is random initialization, and the desired behavior is forced by the fact that the diverse filters (paths) decrease the loss function more compared to the collapsed state. Both of these schemes are similar to ours, and thus, by indirect induction, the threat of collapsing into the same network can be refused. Practically, virtual networks inside VNNE will collapse into the same state only when all virtual networks are initialized equally. The probability of such a state is $(vp^b)^{-1}$, i.e., $2.4\cdot10^{-5}$ considering ResNet, two physical and 20 virtual neural networks


\begin{table}[!h]
      \caption{Performance on CIFAR-10 test dataset using ResNet. The accuracy is the best of the ten runs per model. The prediction time is for a single batch of 50 images. VNNE notation has the format $p/v$. *Because Masksemles performs inference in a different way, we report the inference time of a single image, including necessary preprocessing. The training time denotes time per epoch.}
      \centering
      \resizebox{\textwidth}{!}{
      \begin{tabular}{lrrrrrrrr}
       \hline
	    &Test acc & Params &Mem train& Mem infer&Time train&Time infer&&\\
	    Network&[\%] &[M] &[GB] &[GB]&[s] &[ms]&$d_{dis}$&$d_{KL}$\\
           \hline
           \hline
           ResNet-20& 91.70 &0.27&1.84&1.41&18&4&---&---\\
           ResNet-44 & 91.73 &0.66&1.84&1.41&36&7&---&---\\ 
           ResNet-110 & {92.02} &1.74&2.60&1.41&83&18&---&---\\
           ResNet-164 & 91.58 &2.61&2.60&1.41&125&27&---&---\\
          \hline
        ResNet-20 SWA~\cite{izmailov2018averaging}&{91.35}  &0.27&1.84&1.41&18&4&---&---\\
        ResNet-20 Masksembles~\cite{durasov2021masksembles}&83.15  &0.27&1.84&1.41&23&20*&0.13&0.09\\
        ResNet-44 Masksembles~\cite{durasov2021masksembles}&83.59  &0.66&1.84&1.41&46&43*&0.14&0.12\\
        \hline  
        ResNet-20 ensemble of two&  92.76&0.54&1.84&1.41&36&8&0.09&0.13\\  
        ResNet-20 ensemble of three& 93.19&0.81&1.84&1.41&54&12&0.09&0.08\\    
        ResNet-20 ensemble of four& {93.42}&1.08&1.84&1.41&72&15&0.09&0.07\\    
        \hline
        ResNet-20 2/10 $f_A$ &93.30&0.54&2.60&1.41&121&28&0.09&0.23\\
        ResNet-20 2/20 $f_A$ &{93.40}&0.54&3.64&1.82&230&53&0.08&0.30\\
        \hline
        ResNet-20 2/10 $f_A$ ensemb of two & 93.86&1.08&2.60&1.41&242&57&0.07&0.33\\
        ResNet-20 2/10 $f_A$ ensemb of three & {94.14}&1.62&2.60&1.82&363&85&0.06&0.22\\
        \hline
        ResNet-20 2/20 $f_A$ best single&92.10&0.27&---&1.41&---&4&---&---\\
      \end{tabular}
      }
       \label{table-cifar-resnet}
\end{table}

\begin{table}[!h]
      \caption{Performance on CIFAR-10 test dataset using MobileNetV3. The accuracy is the best of the five runs per model. The prediction time is for a single batch of 50 images. VNNE notation has the format $p/v$.}
      \centering
      \resizebox{\textwidth}{!}{
      \begin{tabular}{lrrrrrrrr}
          \hline
           &Test acc & Params &Mem train& Mem infer&Time train&Time infer&&\\
    	   Network&[\%] &[M] &[GB] &[GB]&[s] &[ms]&$d_{dis}$&$d_{KL}$\\
          \hline
          \hline
          MobileNetV3 small& 70.26 & 1.54 & 4.37 & 4.18 & 19 & 5 & --- & --- \\
          MobileNetV3 large & 75.23 & 4.24 & 7.44 & 7.27 & 23 & 6 & --- & --- \\ 
          \hline
          MobileNetV3 small ensemble of two & 73.28 & 3.08 & 4.37 & 4.18 & 38 & 10 & 0.24 & 0.00 \\
          MobileNetV3 small ensemble of three & 73.35 & 4.62 & 4.37 & 4.18 & 57 & 15 & 0.25 & 0.05 \\
          MobileNetV3 small ensemble of four & 73.15 & 6.16 & 4.37 & 4.18 & 76 & 20 & 0.27 & 0.06 \\
          \hline
          MobileNetV3 small 2/10 $f_A$ & 75.86 & 3.08 & 4.37 & 4.20 & 149 & 30 & 0.64 & 0.20 \\ 
          MobileNetV3 small 2/20 $f_A$ & 80.46 & 3.08 & 4.37 & 4.20 & 282 & 58 & 0.52 & 0.10 \\ 
          \hline
          MobileNetV3 small 2/10 $f_A$ best single & 68.93 & 1.54 & --- & 4.18 & --- & --- & --- & --- \\ 
          MobileNetV3 small 2/20 $f_A$ best single & 75.44 & 1.54 & --- & 4.18 & --- & --- & --- & --- \\ 
          \end{tabular}
      }
       \label{table-cifar-mobilenetv3}
\end{table}  

\begin{table}[!h]
      \caption{Performance on CIFAR-10 test dataset using ConvNextV1. The accuracy is the best of the five runs per model. The prediction time is for a single batch of 50 images. VNNE notation has the format $p/v$.}
      \centering
      \resizebox{\textwidth}{!}{
      \begin{tabular}{lrrrrrrrr}
          \hline
           &Test acc & Params &Mem train& Mem infer&Time train&Time infer&&\\
    	   Network&[\%] &[M] &[GB] &[GB]&[s] &[ms]&$d_{dis}$&$d_{KL}$\\
          \hline
          \hline
          ConvNextV1 atto  & 78.59 & 3.38 & 2.06 & 2.19 & 25 & 4 & --- & --- \\
          ConvNextV1 femto & 78.62 & 4.84 & 2.32 & 2.19 & 27 & 4 & --- & --- \\
          ConvNextV1 pico  & 80.16 & 8.54 & 2.83 & 2.70 & 27 & 4 & --- & --- \\
          ConvNextV1 nano  & 80.89 & 14.96 & 4.37 & 4.18 & 30 & 5 & --- & --- \\
          \hline
          ConvNextV1 atto ensemble of two & 79.90 & 6.76 & 2.06 & 2.19 & 50 & 8 & 0.19 & 0.01 \\
          ConvNextV1 atto ensemble of three & 80.58 & 10.14 & 2.06 & 2.19 & 75 & 12 & 0.19 & 0.00 \\
          ConvNextV1 atto ensemble of four & 81.05 & 13.52 & 2.06 & 2.19 & 100 & 16 & 0.20 & 0.01 \\
          \hline
          ConvNextV1 atto 2/10 $f_A$ & 80.78 & 6.76 & 3.34 & 2.19 & 192 & 24 & 0.56 & 0.08 \\ 
          ConvNextV1 atto 2/20 $f_A$ & 83.61 & 6.76 & 5.39 & 2.19 & 440 & 44 & 0.42 & 0.06 \\ 
          \hline
          ConvNextV1 atto 2/10 $f_A$ best single & 78.91 & 3.38 & --- & 2.19 & --- & --- & --- & --- \\ 
          ConvNextV1 atto 2/20 $f_A$ best single & 81.11 & 3.38 & --- & 2.19 & --- & --- & --- & --- \\ 
          \end{tabular}
      }
       \label{table-cifar-mobilenetv3}
\end{table}  


\begin{table}[!ht]
      \caption{Performance on CIFAR-10 test dataset using EfficientNet. The statistics are given by single run. The prediction time is for a single batch of 16 images. VNNE notation has $p/v$ format. The training time denotes time per epoch.}
      \centering
      \resizebox{\textwidth}{!}{
      \begin{tabular}{lrrrrrrrr}
       \hline
	    &Test acc & Params &Mem train& Mem infer&Time train&Time infer&&\\
	    Network&[\%] &[M] &[GB] &[GB]&[s] &[ms]&$d_{dis}$&$d_{KL}$\\
        \hline
        \hline
        EfficientNet B0&94.97&4.02&5.66&3.35&127&16&---&---\\
        EfficientNet B1&95.04&6.52&9.71&5.38&182&22&---&---\\
        EfficientNet B2&{95.05}&7.71&9.76&5.68&196&26&---&---\\
        \hline
        EfficientNet B0 ensemble of two &95.54&8.04 &5.66 &6.24 &254 &32 &0.04&0.02\\
        EfficientNet B0 ensemble of three &95.88&12.06 &5.66 &9.47 &381 &48 &0.04&0.01\\
        EfficientNet B0 ensemble of four &{95.89}&16.08 &5.66 &12.01 &508 &64 &0.04&0.02\\
        \hline
        EfficientNet B0 2/10 $f_A$&95.72&8.04&34.20&3.35&917&145&0.09&0.22\\ 
        EfficientNet B1 2/10 $f_A$&{96.04}&13.04&34.20&5.38&1420&215&0.08&0.23\\
        \hline
        EfficientNet B0 2/10 $f_A$ best single&95.03&4.02&---&3.35&---&16&---&---\\
        EfficientNet B1 2/10 $f_A$ best single&{95.48}&6.52&---&5.38&---&22&---&---\\
        \hline
      \end{tabular}
      }
      \label{table-cifar-effnet}
\end{table} 

\begin{table}[!h]
      \caption{Performance on CIFAR-100-LT test dataset. The accuracy is the best of the ten runs per model. The prediction time is for a single batch of 50 images. VNNE notation has the format $p/v$. The training time denotes the time per epoch.}
      \centering
      \resizebox{\textwidth}{!}{
      \begin{tabular}{lrrrrrrrr}
       \hline
	    &Test acc & Params &Mem train& Mem infer&Time train&Time infer&&\\
	    Network&[\%] &[M] &[GB] &[GB]&[s] &[ms]&$d_{dis}$&$d_{KL}$\\
           \hline
           \hline
           ResNet-20& 37.23 &0.28&1.84&1.48&5&42&---&---\\
           ResNet-44 &39.28 &0.67&1.94&1.48&9&46&---&---\\ 
           ResNet-110 &39.33 &1.74&2.60&1.49&19&55&---&---\\
           ResNet-164 &39.35 &2.62&2.60&1.49&28&72&---&---\\
        ResNet-20 2/20 $f_A$ &{40.66}&0.56&3.49&1.48&57&86&0.29&0.05\\
        ResNet-20 2/20 $f_A$ best single&39.47&0.28&---&1.48&---&42&---&---\\
        \hline
        ResNet-20 ensemble of two&39.53&0.56&1.84&1.48&10&84&0.52&0.33\\
        ResNet-20 ensemble of three&40.40&0.84&1.84&1.48&15&126&0.52&0.27\\
        ResNet-20 ensemble of four&{41.34}&1.12&1.84&1.48&20&168&0.52&0.30\\
        \hline
        \hline
        EfficientNet B0&42.13  &4.17&11.60&3.40&31&56&---&---\\
        EfficientNet B1&42.48  &6.70&11.60&5.40&43&60&---&---\\
        EfficientNet B2&42.37  &7.90&11.67&5.44&44&71&---&---\\
        EfficientNet B0 2/20 $f_A$&{44.93}  &8.27&34.20&3.91&448&198&0.30&0.06\\
        EfficientNet B0 2/20 $f_A$ best single&43.16  &4.17&---&3.40&---&56&---&---\\
        \hline
        EfficientNet B0 ensemble of two&44.59&8.34&11.60&6.33&62&112&0.44&0.02\\
        EfficientNet B0 ensemble of three&45.28&12.51&11.60&9.96&93&168&0.44&0.05\\
        EfficientNet B0 ensemble of four&{45.99}&16.68&11.60&12.16&124&224&0.44&0.05\\
      \end{tabular}
      }
       \label{table-cifar-resnet100lt}
\end{table}

\subsection{Virtualization of existing architectures}
The major architectures are already split into blocks. Therefore, it is natural to make combinations on the level of blocks instead of layers. This allows us to preserve interconected elements such as batch normalization and its input. Selecting properly the division of architecture into blocks is a challenging task, unless provided explicitly by the architecture itself.

The \emph{ResNet} architecture, regardless of the specific size, is built on repetitive convolutional residual blocks - see Table 1 in \cite{he2016deep}. To avoid affecting the residual connections, each residual block (in the sense of ResNet architecture) is taken as a single block in the meaning of virtual network architecture. However, it is possible to construct a single virtual-network-block from multiple residual-blocks (e.g., for every repetition or the whole same-resolution-set). The input and output layers are taken as specific virtual-network-blocks. In our case, ResNet architectures are split into 11 blocks, giving us 2048 unique combinations of particular virtual networks.

The \emph{EfficientNet} architecture can be divided into the initial block, called \textit{stem} -- performing input rescaling, normalization, padding, convolution, batch normalization, and activation; the middle part, and the final block, called \textit{top} -- performing final convolution, batch normalization, and activation. The middle part of EfficientNet, regardless of the EfficientNet type, consists of seven modules with different complexity with respect to the EfficientNet type. These inner blocks are typically referred as \textit{MBConvX} blocks (see Table 1 in \cite{tan2019efficientnet}). Therefore, our approach considers the stem as one virtual-network-block, every MBConv block as one virtual-network-block, and the top blocks as another virtual-network-block -- nine blocks in total regardless of the EfficientNet type, giving us 512 unique combinations of particular virtual networks.

\section{Benchmark}
\label{sec-benchmark}
\begin{figure}[!h]
    \centering
    \begin{tikzpicture}
    \begin{axis}[height=45mm,width=75mm,legend columns=1, xlabel={Epoch},ylabel={Accuracy},xmin=1, xmax=35, ymin=0.25, ymax=0.85,
    ymajorgrids=true,yminorgrids=true,style={thick},legend style={at={(1.0,0.59)}},axis x line*=bottom,axis y line*=left, style={font=\tiny}]
    \addplot[red,dashed]coordinates{(1,0.073)(2,0.223)(3,0.326)(4,0.395)(5,0.445)(6,0.483)(7,0.514)(8,0.540)(9,0.562)(10,0.581)(11,0.598)(12,0.613)(13,0.626)(14,0.638)(15,0.649)(16,0.659)(17,0.669)(18,0.677)(19,0.685)(20,0.693)(21,0.700)(22,0.707)(23,0.712)(24,0.719)(25,0.725)(26,0.730)(27,0.735)(28,0.740)(29,0.745)(30,0.750)(31,0.774)(32,0.781)(33,0.785)(34,0.788)(35,0.791)};\addlegendentry{EfficientNet B2 train}
    \addplot[red]coordinates{(1,0.160)(2,0.289)(3,0.363)(4,0.418)(5,0.465)(6,0.486)(7,0.514)(8,0.532)(9,0.549)(10,0.564)(11,0.574)(12,0.584)(13,0.598)(14,0.603)(15,0.609)(16,0.609)(17,0.623)(18,0.624)(19,0.624)(20,0.634)(21,0.638)(22,0.636)(23,0.645)(24,0.646)(25,0.650)(26,0.647)(27,0.649)(28,0.651)(29,0.654)(30,0.658)(31,0.664)(32,0.667)(33,0.667)(34,0.667)(35,0.666)};\addlegendentry{EfficientNet B2 valid}
    \addplot[blue,dashed]coordinates{(1,0.080)(2,0.242)(3,0.353)(4,0.428)(5,0.484)(6,0.526)(7,0.559)(8,0.588)(9,0.611)(10,0.631)(11,0.649)(12,0.665)(13,0.679)(14,0.692)(15,0.704)(16,0.715)(17,0.725)(18,0.735)(19,0.744)(20,0.752)(21,0.760)(22,0.767)(23,0.774)(24,0.780)(25,0.787)(26,0.793)(27,0.799)(28,0.805)(29,0.810)(30,0.815)(31,0.836)(32,0.842)(33,0.846)(34,0.849)(35,0.852)};\addlegendentry{EfficientNet B2 2/10 train}
    \addplot[blue]coordinates{(1,0.152)(2,0.304)(3,0.393)(4,0.447)(5,0.488)(6,0.518)(7,0.544)(8,0.565)(9,0.574)(10,0.591)(11,0.602)(12,0.608)(13,0.622)(14,0.621)(15,0.631)(16,0.639)(17,0.636)(18,0.642)(19,0.653)(20,0.651)(21,0.657)(22,0.659)(23,0.658)(24,0.661)(25,0.664)(26,0.666)(27,0.668)(28,0.671)(29,0.673)(30,0.673)(31,0.685)(32,0.686)(33,0.682)(34,0.683)(35,0.687)};\addlegendentry{EfficientNet B2 2/10 valid}
    \end{axis}
    \end{tikzpicture}
    \caption{EfficientNet B2 and EfficientNet B2 2/10 performance on ImageNet during 35 epochs.}
    \label{fig-graph-imagenet}
\end{figure}
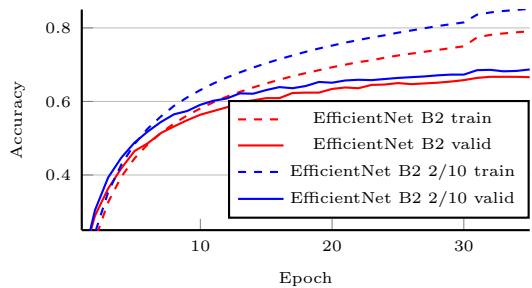

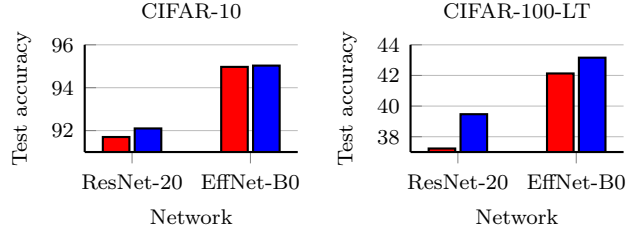
\begin{figure}[!h]
    \centering
    \begin{subfigure}[b]{0.23\textwidth}
    \begin{tikzpicture}
    \begin{axis}
        [
        title=CIFAR-10,
        height=30mm,
        width=44mm,
        ybar,
        xtick=data,
        bar width=10pt,
        enlarge x limits=0.4,
        xlabel={Network},
        ylabel={Test accuracy},
        ymin=91,
        ymax=96,
        ymajorgrids=true,
        yminorgrids=true,
        style={thick},
        axis x line*=bottom,
        axis y line*=left,
        style={font=\scriptsize},
        symbolic x coords={ResNet-20, EffNet-B0}
        ]
    \addplot[fill=red]coordinates{(ResNet-20,91.7)(EffNet-B0,94.97)};
    \addplot[fill=blue]coordinates{(ResNet-20,92.1)(EffNet-B0,95.03)};
    \legend{}
    \end{axis}
    \end{tikzpicture}
    \label{graph-ablation-b}
    \end{subfigure}
    ~~
    \begin{subfigure}[b]{0.23\textwidth}
    \begin{tikzpicture}
    \begin{axis}
        [
        title=CIFAR-100-LT,
        height=30mm,
        width=44mm,
        ybar,
        xtick=data,
        bar width=10pt,
        enlarge x limits=0.4,
        xlabel={Network},
        ylabel={Test accuracy},
        ymin=37,
        ymax=44,
        ymajorgrids=true,
        yminorgrids=true,
        style={thick},
        axis x line*=bottom,
        axis y line*=left,
        style={font=\scriptsize},
        symbolic x coords={ResNet-20, EffNet-B0}
        ]
    \addplot[fill=red]coordinates{(ResNet-20,37.23)(EffNet-B0,42.13)};
    \addplot[fill=blue]coordinates{(ResNet-20,39.47)(EffNet-B0,43.16)};
    \end{axis}
    \end{tikzpicture}
    \label{graph-ablation-b}
    \end{subfigure}
    \caption{Comparison of standalone trained networks (red bars) and the best virtual network extracted from VNNE (blue bars). }
    \label{gprahs-ablations}
\end{figure}

We involve three datasets: CIFAR-10~\cite{krizhevsky2009learning} (Tables~\ref{table-cifar-resnet}, \ref{table-cifar-effnet} and Fig.~\ref{gprahs-ablations}), CIFAR-100-LT~\cite{cao2019learning} (Table~\ref{table-cifar-resnet100lt} and Fig.~\ref{gprahs-ablations}), and ImageNet~\cite{ILSVRC15} (Fig.~\ref{fig-graph-imagenet}) in which various settings of VNNE are trained. The goal of the benchmark is not to establish a new SOTA or to perfectly replicate the best reported results, but to compare the baseline with VNNE for the same training settings.

\textbf{CIFAR-10:} We use the standard train/test split. As a baseline, we used ResNet~\cite{he2016deep} and EfficientNet~\cite{tan2019efficientnet}, where for EfficientNet we upscaled the images to the native resolution of the models (see \cite{tan2019efficientnet}). The models are trained for 200 epochs, batch size 50 (ResNet models) or 16 (EfficientNet models), Adam optimizer~\cite{kingma2014adam} with $\alpha=0.001$ and its decrease by factor 10 at epochs 120, 160, and 185. VNNEs are constructed with $\omega=0.15$ which was determined by the Optuna~\cite{akiba2019optuna} hyperparameter search, see more in Section~\ref{sec-ablation_study}. Masksembles ResNet~\cite{durasov2021masksembles} is trained for five virtual networks and overlapped with degree 2. As data augmentation, we use shifts and flips. The training was performed on the A-100 and inference on RTX-3090 graphics cards. To evaluate the diversity of the models in the ensemble, we measured the diversity disagreement ($d_{dis}$) and the Kullback-Leibler diversity ($d_{KL}$); for detailed definitions, see~\cite{liu2021deep}. The greater the diversity, the better. For single VNNE (for both 2/10 and 2/20 settings), we measured the divergence of inner virtual networks; for their ensemble, we measured the divergence of aggregated outputs. We also evaluated particular virtual networks in VNNE and selected the best one. In the tables, it is notated as the ``best single''. We can take a single virtual network from the ensemble and use it standalone as a standard model owing to the utilization of a simple aggregation layer without trainable parameters.

\textbf{CIFAR-100-LT:} It is a modified version of standard CIFAR-100 where predefined samples are deleted to create a long-tailed distribution~\cite{cao2019learning} of the data class. Such a task is much more complicated as it can include, e.g., five samples per class in the train set, but 500 samples in the test set, resulting in easy-achieved overfit. The dataset has been created with an imbalance ratio of 100 as described in~\cite{cao2019learning}. The training setting is identical to the CIFAR-10 setting.

\textbf{ImageNet:} We have used the standard train/valid split, fixed input resolution of $260^2$px, batch size 48, and trained it for 35 epochs with Adam optimizer with $\alpha=1e^{-4}$ and decreased to $\alpha=5e^{-5}$ after 30$^{th}$ epochs. The augmentations are shifts, flips and resizes. The models are EfficientNetB2 and EfficientNetB2 2/10 with $\omega=0.15$ and $f_A$ as aggregation layers. Evaluation of the virtual neural network ensemble in the ImageNet dataset~\cite{ILSVRC15} is shown in Figure~\ref{fig-graph-imagenet} with the best
 validation top-1 accuracy of 0.667 for vanilla and 0.687 for VNNE.

\noindent\textbf{The overall results show the following:}
\begin{itemize}
    \item VNNE reaches a higher test accuracy than the baseline. The accuracy of a VNNE is higher than that of a heavier baseline with more trainable parameters than a VNNE has.
    \item The best single particular virtual network extracted from VNNE has higher accuracy than the baseline and even higher accuracy than the heavier baselines that have more trainable parameters.
    \item To obtain an accuracy comparable to that of a VNNE with two physical networks, a standard deep ensemble of four models must be established. Moreover, VNNEs can be ensembled as well. 
    \item Compared to Masksembles and standard deep ensembles, VNNEs have a lower disagreement diversity and a higher Kullback-Leibler diversity. Taking into account the precision, it means that VNNEs are less over-confident in wrong predictions than the other approaches.
\end{itemize}
This leads to the final conclusion. \textbf{VNNEs are suitable for applications where accuracy is crucial and where model size is more important than processing speed.}

\section{Ablation study}
\label{sec-ablation_study}
\textbf{Scalability.} The graph in Figure~\ref{fig-graph-scale-vritual} presents a key result: the scalability of virtual capacity. The number of physical networks was fixed to two ResNet-20 models, and VNNEs were trained with varying numbers of virtual networks (ranging from 2 to 30) for 80 epochs on the CIFAR-10 dataset. The test accuracy was then measured to evaluate performance. In the graph, the more red the color, the more virtual networks are used. All VNNEs have the same number of trainable parameters. The correlation between the number of virtual models and the test accuracy is visually obvious and is numerically 0.95 for [2, 10] virtual networks, 0.80 for [2, 20], and 0.62 for [2, 30]. This differs from most approaches, where there is a correlation between the number of trainable parameters and the accuracy~\cite{dehghani2021efficiency}. In our case, the number of trainable parameters is fixed. Note that all trainings were performed with identical $\omega$ and learning rate, so no further adjustment of hyperparameters was made.

\definecolor{c1}{RGB}{0,0,97}
\definecolor{c2}{RGB}{0,0,170}
\definecolor{c3}{RGB}{0,0,219}
\definecolor{c4}{RGB}{0,44,219}
\definecolor{c5}{RGB}{0,109,219}
\definecolor{c6}{RGB}{0,174,219}
\definecolor{c7}{RGB}{10,219,174}
\definecolor{c8}{RGB}{63,219,122}
\definecolor{c9}{RGB}{114,219,72}
\definecolor{c10}{RGB}{165,219,20}
\definecolor{c11}{RGB}{218,202,0}
\definecolor{c12}{RGB}{219,141,0}
\definecolor{c13}{RGB}{219,81,0}
\definecolor{c14}{RGB}{219,20,0}
\definecolor{c15}{RGB}{178,0,0}

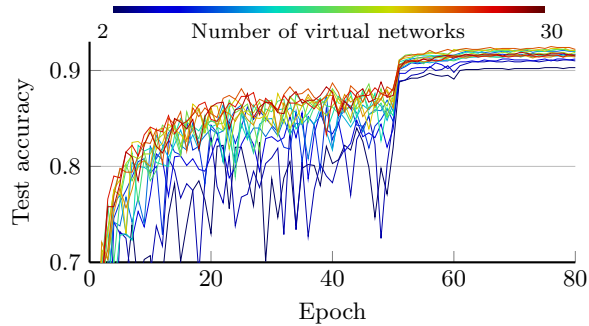
\begin{figure}[!h]
    \centering
    ~~~~~~\includegraphics[width=57mm]{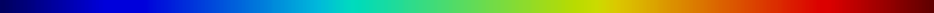}\\
    \scriptsize{~~~~~~~~2~~~~~~~~~~~Number of virtual networks~~~~~~~~~~30}\\
    \begin{tikzpicture}
    \begin{axis}[height=45mm,width=80mm,legend columns=1, xlabel={Epoch},ylabel={Test accuracy},xmin=0, xmax=80, ymin=0.7, ymax=0.93,
    ymajorgrids=true,yminorgrids=true,style={thick},legend style={at={(1.0,0.31)}},axis x line*=bottom,axis y line*=left, style={font=\footnotesize}]
    \addplot[color=c1, line width=0.1mm]coordinates{(1,0.540)(2,0.553)(3,0.683)(4,0.688)(5,0.655)(6,0.680)(7,0.739)(8,0.675)(9,0.703)(10,0.707)(11,0.624)(12,0.685)(13,0.749)(14,0.782)(15,0.701)(16,0.739)(17,0.775)(18,0.785)(19,0.765)(20,0.730)(21,0.821)(22,0.822)(23,0.756)(24,0.820)(25,0.820)(26,0.791)(27,0.759)(28,0.825)(29,0.701)(30,0.787)(31,0.747)(32,0.788)(33,0.805)(34,0.738)(35,0.798)(36,0.802)(37,0.801)(38,0.780)(39,0.821)(40,0.803)(41,0.807)(42,0.822)(43,0.817)(44,0.840)(45,0.826)(46,0.799)(47,0.747)(48,0.798)(49,0.771)(50,0.820)(51,0.888)(52,0.890)(53,0.892)(54,0.892)(55,0.894)(56,0.897)(57,0.895)(58,0.894)(59,0.895)(60,0.891)(61,0.900)(62,0.901)(63,0.901)(64,0.901)(65,0.902)(66,0.902)(67,0.902)(68,0.902)(69,0.901)(70,0.902)(71,0.902)(72,0.902)(73,0.902)(74,0.902)(75,0.903)(76,0.903)(77,0.903)(78,0.902)(79,0.903)(80,0.903)};
    \addplot[color=c2, line width=0.1mm]coordinates{(1,0.430)(2,0.615)(3,0.650)(4,0.580)(5,0.732)(6,0.725)(7,0.675)(8,0.668)(9,0.661)(10,0.688)(11,0.738)(12,0.717)(13,0.808)(14,0.796)(15,0.767)(16,0.782)(17,0.775)(18,0.688)(19,0.786)(20,0.830)(21,0.779)(22,0.762)(23,0.811)(24,0.738)(25,0.768)(26,0.771)(27,0.795)(28,0.778)(29,0.803)(30,0.809)(31,0.819)(32,0.729)(33,0.839)(34,0.734)(35,0.805)(36,0.727)(37,0.776)(38,0.802)(39,0.790)(40,0.783)(41,0.806)(42,0.818)(43,0.770)(44,0.828)(45,0.841)(46,0.852)(47,0.823)(48,0.725)(49,0.802)(50,0.847)(51,0.889)(52,0.890)(53,0.895)(54,0.896)(55,0.898)(56,0.904)(57,0.900)(58,0.901)(59,0.905)(60,0.898)(61,0.905)(62,0.907)(63,0.909)(64,0.909)(65,0.909)(66,0.910)(67,0.909)(68,0.909)(69,0.910)(70,0.909)(71,0.909)(72,0.910)(73,0.910)(74,0.909)(75,0.910)(76,0.911)(77,0.910)(78,0.911)(79,0.911)(80,0.910)};
    \addplot[color=c3, line width=0.1mm]coordinates{(1,0.567)(2,0.645)(3,0.559)(4,0.742)(5,0.730)(6,0.729)(7,0.724)(8,0.753)(9,0.737)(10,0.770)(11,0.813)(12,0.807)(13,0.769)(14,0.810)(15,0.811)(16,0.801)(17,0.808)(18,0.797)(19,0.795)(20,0.811)(21,0.829)(22,0.818)(23,0.788)(24,0.799)(25,0.823)(26,0.835)(27,0.823)(28,0.795)(29,0.829)(30,0.831)(31,0.826)(32,0.830)(33,0.852)(34,0.830)(35,0.856)(36,0.787)(37,0.834)(38,0.856)(39,0.838)(40,0.834)(41,0.846)(42,0.816)(43,0.824)(44,0.848)(45,0.831)(46,0.835)(47,0.830)(48,0.844)(49,0.820)(50,0.839)(51,0.897)(52,0.897)(53,0.902)(54,0.903)(55,0.903)(56,0.903)(57,0.908)(58,0.906)(59,0.902)(60,0.906)(61,0.909)(62,0.909)(63,0.910)(64,0.909)(65,0.911)(66,0.910)(67,0.910)(68,0.912)(69,0.912)(70,0.911)(71,0.911)(72,0.911)(73,0.910)(74,0.911)(75,0.912)(76,0.910)(77,0.912)(78,0.912)(79,0.911)(80,0.912)};
    \addplot[color=c4,line width=0.1mm]coordinates{(1,0.496)(2,0.665)(3,0.662)(4,0.746)(5,0.763)(6,0.775)(7,0.789)(8,0.805)(9,0.808)(10,0.793)(11,0.800)(12,0.729)(13,0.828)(14,0.815)(15,0.812)(16,0.847)(17,0.817)(18,0.821)(19,0.782)(20,0.843)(21,0.835)(22,0.805)(23,0.846)(24,0.824)(25,0.841)(26,0.862)(27,0.849)(28,0.858)(29,0.850)(30,0.843)(31,0.836)(32,0.820)(33,0.818)(34,0.857)(35,0.805)(36,0.852)(37,0.843)(38,0.796)(39,0.861)(40,0.828)(41,0.844)(42,0.868)(43,0.862)(44,0.843)(45,0.842)(46,0.819)(47,0.842)(48,0.849)(49,0.860)(50,0.847)(51,0.901)(52,0.904)(53,0.908)(54,0.907)(55,0.908)(56,0.910)(57,0.911)(58,0.910)(59,0.910)(60,0.913)(61,0.915)(62,0.916)(63,0.916)(64,0.915)(65,0.915)(66,0.916)(67,0.916)(68,0.917)(69,0.916)(70,0.916)(71,0.917)(72,0.916)(73,0.918)(74,0.916)(75,0.917)(76,0.916)(77,0.916)(78,0.917)(79,0.916)(80,0.917)};
    \addplot[color=c5, line width=0.1mm]coordinates{(1,0.543)(2,0.681)(3,0.678)(4,0.742)(5,0.738)(6,0.768)(7,0.742)(8,0.767)(9,0.741)(10,0.821)(11,0.808)(12,0.819)(13,0.835)(14,0.843)(15,0.822)(16,0.817)(17,0.845)(18,0.824)(19,0.832)(20,0.839)(21,0.834)(22,0.837)(23,0.848)(24,0.859)(25,0.864)(26,0.829)(27,0.840)(28,0.840)(29,0.831)(30,0.841)(31,0.832)(32,0.849)(33,0.833)(34,0.845)(35,0.844)(36,0.808)(37,0.845)(38,0.852)(39,0.874)(40,0.819)(41,0.851)(42,0.858)(43,0.863)(44,0.856)(45,0.836)(46,0.854)(47,0.859)(48,0.858)(49,0.869)(50,0.858)(51,0.901)(52,0.909)(53,0.912)(54,0.913)(55,0.914)(56,0.914)(57,0.915)(58,0.915)(59,0.916)(60,0.916)(61,0.918)(62,0.920)(63,0.920)(64,0.919)(65,0.919)(66,0.918)(67,0.919)(68,0.920)(69,0.920)(70,0.921)(71,0.921)(72,0.921)(73,0.919)(74,0.920)(75,0.920)(76,0.920)(77,0.920)(78,0.921)(79,0.920)(80,0.920)};
    \addplot[color=c6, line width=0.1mm]coordinates{(1,0.584)(2,0.676)(3,0.698)(4,0.761)(5,0.770)(6,0.730)(7,0.804)(8,0.800)(9,0.799)(10,0.795)(11,0.832)(12,0.822)(13,0.809)(14,0.823)(15,0.838)(16,0.819)(17,0.833)(18,0.817)(19,0.845)(20,0.847)(21,0.847)(22,0.864)(23,0.829)(24,0.804)(25,0.855)(26,0.836)(27,0.814)(28,0.844)(29,0.849)(30,0.850)(31,0.836)(32,0.855)(33,0.861)(34,0.856)(35,0.844)(36,0.851)(37,0.844)(38,0.861)(39,0.836)(40,0.817)(41,0.858)(42,0.850)(43,0.841)(44,0.852)(45,0.864)(46,0.859)(47,0.868)(48,0.863)(49,0.848)(50,0.866)(51,0.902)(52,0.905)(53,0.909)(54,0.908)(55,0.909)(56,0.908)(57,0.909)(58,0.910)(59,0.911)(60,0.911)(61,0.913)(62,0.913)(63,0.914)(64,0.914)(65,0.916)(66,0.915)(67,0.914)(68,0.914)(69,0.915)(70,0.915)(71,0.915)(72,0.916)(73,0.914)(74,0.915)(75,0.916)(76,0.915)(77,0.915)(78,0.916)(79,0.915)(80,0.915)};
    \addplot[color=c7, line width=0.1mm]coordinates{(1,0.538)(2,0.692)(3,0.734)(4,0.668)(5,0.728)(6,0.730)(7,0.723)(8,0.822)(9,0.816)(10,0.819)(11,0.804)(12,0.802)(13,0.800)(14,0.832)(15,0.850)(16,0.840)(17,0.842)(18,0.837)(19,0.848)(20,0.841)(21,0.844)(22,0.805)(23,0.863)(24,0.808)(25,0.844)(26,0.828)(27,0.864)(28,0.854)(29,0.869)(30,0.862)(31,0.859)(32,0.860)(33,0.867)(34,0.839)(35,0.849)(36,0.855)(37,0.865)(38,0.871)(39,0.836)(40,0.854)(41,0.873)(42,0.867)(43,0.877)(44,0.851)(45,0.876)(46,0.860)(47,0.876)(48,0.827)(49,0.861)(50,0.853)(51,0.909)(52,0.911)(53,0.913)(54,0.916)(55,0.916)(56,0.916)(57,0.915)(58,0.916)(59,0.914)(60,0.914)(61,0.917)(62,0.919)(63,0.919)(64,0.918)(65,0.921)(66,0.920)(67,0.920)(68,0.921)(69,0.918)(70,0.921)(71,0.920)(72,0.919)(73,0.921)(74,0.920)(75,0.920)(76,0.921)(77,0.922)(78,0.922)(79,0.921)(80,0.920)};
    \addplot[color=c8, line width=0.1mm]coordinates{(1,0.572)(2,0.696)(3,0.760)(4,0.700)(5,0.765)(6,0.797)(7,0.814)(8,0.763)(9,0.808)(10,0.785)(11,0.796)(12,0.818)(13,0.832)(14,0.841)(15,0.814)(16,0.847)(17,0.839)(18,0.829)(19,0.853)(20,0.822)(21,0.841)(22,0.837)(23,0.844)(24,0.789)(25,0.857)(26,0.846)(27,0.815)(28,0.839)(29,0.860)(30,0.857)(31,0.868)(32,0.870)(33,0.834)(34,0.875)(35,0.867)(36,0.846)(37,0.862)(38,0.871)(39,0.863)(40,0.862)(41,0.855)(42,0.859)(43,0.846)(44,0.864)(45,0.879)(46,0.854)(47,0.857)(48,0.866)(49,0.851)(50,0.859)(51,0.906)(52,0.910)(53,0.908)(54,0.911)(55,0.907)(56,0.912)(57,0.914)(58,0.913)(59,0.911)(60,0.910)(61,0.912)(62,0.915)(63,0.914)(64,0.916)(65,0.915)(66,0.914)(67,0.914)(68,0.914)(69,0.915)(70,0.916)(71,0.914)(72,0.916)(73,0.915)(74,0.915)(75,0.917)(76,0.916)(77,0.916)(78,0.915)(79,0.915)(80,0.917)};
    \addplot[color=c9, line width=0.1mm]coordinates{(1,0.412)(2,0.650)(3,0.735)(4,0.774)(5,0.796)(6,0.802)(7,0.785)(8,0.799)(9,0.815)(10,0.790)(11,0.819)(12,0.843)(13,0.838)(14,0.833)(15,0.811)(16,0.857)(17,0.837)(18,0.822)(19,0.841)(20,0.848)(21,0.844)(22,0.845)(23,0.869)(24,0.856)(25,0.860)(26,0.865)(27,0.844)(28,0.816)(29,0.868)(30,0.869)(31,0.832)(32,0.857)(33,0.873)(34,0.865)(35,0.863)(36,0.868)(37,0.862)(38,0.865)(39,0.869)(40,0.881)(41,0.871)(42,0.860)(43,0.862)(44,0.870)(45,0.874)(46,0.886)(47,0.861)(48,0.881)(49,0.853)(50,0.873)(51,0.911)(52,0.916)(53,0.917)(54,0.915)(55,0.917)(56,0.919)(57,0.918)(58,0.918)(59,0.917)(60,0.921)(61,0.922)(62,0.921)(63,0.921)(64,0.922)(65,0.921)(66,0.922)(67,0.921)(68,0.921)(69,0.921)(70,0.920)(71,0.922)(72,0.922)(73,0.921)(74,0.921)(75,0.922)(76,0.921)(77,0.922)(78,0.922)(79,0.921)(80,0.921)};
    \addplot[color=c10, line width=0.1mm]coordinates{(1,0.552)(2,0.713)(3,0.703)(4,0.775)(5,0.770)(6,0.803)(7,0.811)(8,0.808)(9,0.813)(10,0.813)(11,0.827)(12,0.833)(13,0.840)(14,0.844)(15,0.821)(16,0.846)(17,0.854)(18,0.860)(19,0.835)(20,0.858)(21,0.843)(22,0.859)(23,0.818)(24,0.850)(25,0.856)(26,0.858)(27,0.856)(28,0.867)(29,0.847)(30,0.865)(31,0.859)(32,0.845)(33,0.872)(34,0.867)(35,0.845)(36,0.865)(37,0.869)(38,0.860)(39,0.853)(40,0.863)(41,0.871)(42,0.861)(43,0.873)(44,0.860)(45,0.868)(46,0.881)(47,0.871)(48,0.870)(49,0.863)(50,0.882)(51,0.904)(52,0.909)(53,0.910)(54,0.913)(55,0.913)(56,0.915)(57,0.913)(58,0.914)(59,0.913)(60,0.916)(61,0.916)(62,0.916)(63,0.918)(64,0.916)(65,0.918)(66,0.916)(67,0.917)(68,0.919)(69,0.918)(70,0.918)(71,0.915)(72,0.918)(73,0.917)(74,0.918)(75,0.917)(76,0.919)(77,0.918)(78,0.919)(79,0.917)(80,0.919)};
    \addplot[color=c11, line width=0.1mm]coordinates{(1,0.596)(2,0.705)(3,0.731)(4,0.776)(5,0.786)(6,0.730)(7,0.825)(8,0.826)(9,0.821)(10,0.840)(11,0.818)(12,0.824)(13,0.844)(14,0.839)(15,0.833)(16,0.826)(17,0.865)(18,0.859)(19,0.841)(20,0.863)(21,0.863)(22,0.839)(23,0.861)(24,0.870)(25,0.864)(26,0.855)(27,0.851)(28,0.850)(29,0.853)(30,0.842)(31,0.876)(32,0.862)(33,0.840)(34,0.850)(35,0.863)(36,0.872)(37,0.864)(38,0.858)(39,0.868)(40,0.879)(41,0.866)(42,0.865)(43,0.850)(44,0.874)(45,0.860)(46,0.839)(47,0.867)(48,0.871)(49,0.880)(50,0.867)(51,0.911)(52,0.913)(53,0.914)(54,0.917)(55,0.917)(56,0.918)(57,0.918)(58,0.917)(59,0.916)(60,0.922)(61,0.920)(62,0.921)(63,0.920)(64,0.921)(65,0.921)(66,0.920)(67,0.923)(68,0.922)(69,0.922)(70,0.920)(71,0.921)(72,0.920)(73,0.921)(74,0.922)(75,0.920)(76,0.923)(77,0.923)(78,0.924)(79,0.921)(80,0.921)};
    \addplot[color=c12, line width=0.1mm]coordinates{(1,0.580)(2,0.721)(3,0.703)(4,0.752)(5,0.777)(6,0.757)(7,0.776)(8,0.802)(9,0.806)(10,0.800)(11,0.837)(12,0.840)(13,0.814)(14,0.849)(15,0.848)(16,0.828)(17,0.838)(18,0.860)(19,0.840)(20,0.861)(21,0.841)(22,0.858)(23,0.855)(24,0.847)(25,0.859)(26,0.849)(27,0.843)(28,0.860)(29,0.854)(30,0.868)(31,0.836)(32,0.866)(33,0.852)(34,0.866)(35,0.863)(36,0.875)(37,0.866)(38,0.868)(39,0.874)(40,0.874)(41,0.858)(42,0.875)(43,0.867)(44,0.872)(45,0.852)(46,0.861)(47,0.865)(48,0.876)(49,0.878)(50,0.861)(51,0.902)(52,0.907)(53,0.910)(54,0.911)(55,0.909)(56,0.910)(57,0.910)(58,0.911)(59,0.913)(60,0.915)(61,0.914)(62,0.913)(63,0.915)(64,0.915)(65,0.915)(66,0.915)(67,0.916)(68,0.915)(69,0.915)(70,0.916)(71,0.916)(72,0.915)(73,0.913)(74,0.915)(75,0.914)(76,0.914)(77,0.916)(78,0.914)(79,0.915)(80,0.914)};
    \addplot[color=c13, line width=0.1mm]coordinates{(1,0.592)(2,0.691)(3,0.713)(4,0.769)(5,0.798)(6,0.794)(7,0.794)(8,0.811)(9,0.825)(10,0.824)(11,0.839)(12,0.803)(13,0.850)(14,0.844)(15,0.857)(16,0.816)(17,0.861)(18,0.852)(19,0.861)(20,0.855)(21,0.851)(22,0.871)(23,0.874)(24,0.870)(25,0.881)(26,0.870)(27,0.862)(28,0.870)(29,0.867)(30,0.864)(31,0.857)(32,0.875)(33,0.883)(34,0.875)(35,0.876)(36,0.881)(37,0.872)(38,0.868)(39,0.873)(40,0.887)(41,0.877)(42,0.857)(43,0.884)(44,0.878)(45,0.867)(46,0.880)(47,0.863)(48,0.870)(49,0.883)(50,0.879)(51,0.910)(52,0.915)(53,0.916)(54,0.917)(55,0.917)(56,0.921)(57,0.918)(58,0.920)(59,0.922)(60,0.920)(61,0.922)(62,0.923)(63,0.922)(64,0.922)(65,0.923)(66,0.924)(67,0.922)(68,0.923)(69,0.923)(70,0.923)(71,0.922)(72,0.923)(73,0.923)(74,0.924)(75,0.923)(76,0.925)(77,0.924)(78,0.924)(79,0.924)(80,0.922)};
    \addplot[color=c14, line width=0.1mm]coordinates{(1,0.615)(2,0.675)(3,0.766)(4,0.790)(5,0.788)(6,0.815)(7,0.822)(8,0.811)(9,0.833)(10,0.841)(11,0.838)(12,0.836)(13,0.853)(14,0.834)(15,0.843)(16,0.847)(17,0.854)(18,0.864)(19,0.867)(20,0.849)(21,0.847)(22,0.856)(23,0.866)(24,0.862)(25,0.873)(26,0.873)(27,0.877)(28,0.854)(29,0.871)(30,0.865)(31,0.859)(32,0.874)(33,0.866)(34,0.866)(35,0.870)(36,0.865)(37,0.860)(38,0.869)(39,0.833)(40,0.872)(41,0.855)(42,0.871)(43,0.879)(44,0.872)(45,0.874)(46,0.874)(47,0.882)(48,0.876)(49,0.873)(50,0.881)(51,0.906)(52,0.908)(53,0.910)(54,0.910)(55,0.912)(56,0.912)(57,0.910)(58,0.911)(59,0.915)(60,0.913)(61,0.915)(62,0.915)(63,0.916)(64,0.914)(65,0.915)(66,0.915)(67,0.916)(68,0.915)(69,0.914)(70,0.915)(71,0.915)(72,0.916)(73,0.915)(74,0.914)(75,0.915)(76,0.916)(77,0.916)(78,0.914)(79,0.917)(80,0.915)};
    \addplot[color=c15, line width=0.1mm]coordinates{(1,0.569)(2,0.668)(3,0.758)(4,0.758)(5,0.782)(6,0.815)(7,0.810)(8,0.785)(9,0.823)(10,0.836)(11,0.821)(12,0.814)(13,0.827)(14,0.850)(15,0.846)(16,0.839)(17,0.849)(18,0.818)(19,0.852)(20,0.862)(21,0.866)(22,0.867)(23,0.839)(24,0.859)(25,0.861)(26,0.876)(27,0.851)(28,0.862)(29,0.867)(30,0.866)(31,0.856)(32,0.855)(33,0.872)(34,0.878)(35,0.866)(36,0.869)(37,0.869)(38,0.874)(39,0.883)(40,0.877)(41,0.853)(42,0.868)(43,0.860)(44,0.876)(45,0.881)(46,0.867)(47,0.870)(48,0.888)(49,0.872)(50,0.874)(51,0.902)(52,0.910)(53,0.910)(54,0.909)(55,0.912)(56,0.910)(57,0.914)(58,0.915)(59,0.916)(60,0.915)(61,0.915)(62,0.917)(63,0.916)(64,0.916)(65,0.917)(66,0.917)(67,0.917)(68,0.915)(69,0.917)(70,0.917)(71,0.919)(72,0.918)(73,0.916)(74,0.917)(75,0.916)(76,0.918)(77,0.918)(78,0.918)(79,0.916)(80,0.916)};
    \end{axis}
    \end{tikzpicture}

    \caption{Scalability of virtual capacity. The more red the color, the higher the number of virtual networks in the VNNE.}
    \label{fig-graph-scale-vritual}
\end{figure}

\textbf{Setting of $\mathbf{\omega}$.} We used Optuna for the hyperparameter search of $\omega$ with the finding that $\omega \approx 0.15$ produces the best score. The finding holds for both VNNE 2/10 and 2/20. There is only a negligible decrease in precision for $\omega \in [0.1, 0.2]$. On the other hand, when $\omega > 0.4$, the accuracy drops by more than 10\% down. In such a setting, the particular networks do not fit together in the ensemble.

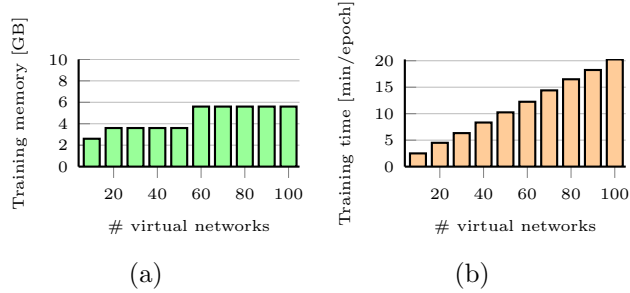
\begin{figure}[!h]
    \centering
    \begin{subfigure}[b]{0.23\textwidth}
        \begin{tikzpicture}
        \begin{axis}[height=30mm,width=45mm,ybar,bar width=6pt,legend columns=1, xlabel={$\#$ virtual networks},ylabel={Training memory [GB]},xmin=4, xmax=105, ymin=0, ymax=10,
        ymajorgrids=true,yminorgrids=true,style={thick},legend style={at={(0.2,0.6)}},axis x line*=bottom,axis y line*=left, style={font=\tiny}]
        \addplot[fill=green!40!white]coordinates{(10,2.6)(20,3.6)(30,3.6)(40,3.6)(50,3.6)(60,5.6)(70,5.6)(80,5.6)(90,5.6)(100,5.6)};
        \end{axis}
        \end{tikzpicture}
        \caption{}
        \label{graph-ablation-b}
    \end{subfigure}
    ~~
    \begin{subfigure}[b]{0.23\textwidth}
        \begin{tikzpicture}
        \begin{axis}[height=30mm,width=45mm,ybar,bar width=6pt,legend columns=1, xlabel={$\#$ virtual networks},ylabel={Training time [min/epoch]},xmin=4, xmax=105, ymin=0, ymax=20.25,
        ymajorgrids=true,yminorgrids=true,style={thick},legend style={at={(0.2,0.6)}},axis x line*=bottom,axis y line*=left, style={font=\tiny}]
        \addplot[fill=orange!40!white]coordinates{(10,2.5)(20,4.5)(30,6.33)(40,8.33)(50,10.25)(60,12.25)(70,14.4)(80,16.5)(90,18.25)(100,20.25)};
        \end{axis}
        \end{tikzpicture}
        \caption{}
        \label{graph-ablation-c}
    \end{subfigure}
    \caption{(a): GPU memory complexity with increasing number of virtual networks; (b) time complexity with increasing number of virtual networks.}
\end{figure}

\textbf{Memory and computational complexity.}
According to Graph~\ref{graph-ablation-b}, the measured memory consumption during training is sublinear and remains constant piecewise, likely due to internal optimizations in TensorFlow. Simultaneous training of two physical networks, combined into 100 virtual neural networks, requires only approximately $4\times$ more memory than training a single standard model.  
The training time, presented in Graph~\ref{graph-ablation-c}, increases linearly with the number of virtual networks involved. Consequently, by scaling the virtual capacity, both accuracy (Table~\ref{table-cifar-resnet}) and computational complexity can be controlled while keeping the number of trainable parameters constant. This contrasts with most networks, where prior studies~\cite{dehghani2021efficiency} have shown that the primary factor influencing accuracy is the number of trainable parameters in a model.

Comparing VNNE with traditional ensembling (and its variants), both require the same computational resources for training the same number of (sub) models, as the number of multiplications/addings is identical. The only and main difference is the memory consumption. We illustrate in Graph~\ref{graph-ablation-b} that the VNNE memory requirement grows sublinearly because the number of trainable weights is constant. However, simultaneous training of n networks in the standard ensemble requires n times m of memory. This is the reason why we did not train n models of the standard ensemble simultaneously (and trained them one-by-one), they cannot fit to memory.

\begin{figure}
    \centering
    \begin{tikzpicture}
    \begin{axis}[height=40mm,width=70mm,legend columns=3, xlabel={Epoch},ylabel={Accuracy},xmin=1, xmax=60, ymin=0.65, ymax=0.92,
    ymajorgrids=true,yminorgrids=true,style={thick},legend style={at={(1.01,1.35)}},axis x line*=bottom,axis y line*=left, style={font=\tiny}]
    \addplot[black,dashed]coordinates{(1,0.493)(2,0.649)(3,0.707)(4,0.740)(5,0.762)(6,0.778)(7,0.791)(8,0.799)(9,0.805)(10,0.813)(11,0.819)(12,0.826)(13,0.828)(14,0.834)(15,0.836)(16,0.836)(17,0.841)(18,0.842)(19,0.845)(20,0.847)(21,0.849)(22,0.852)(23,0.855)(24,0.855)(25,0.857)(26,0.858)(27,0.857)(28,0.860)(29,0.859)(30,0.860)(31,0.861)(32,0.862)(33,0.863)(34,0.865)(35,0.866)(36,0.865)(37,0.867)(38,0.867)(39,0.868)(40,0.867)(41,0.868)(42,0.870)(43,0.868)(44,0.871)(45,0.870)(46,0.871)(47,0.872)(48,0.872)(49,0.872)(50,0.874)(51,0.877)(52,0.873)(53,0.876)(54,0.876)(55,0.876)(56,0.877)(57,0.876)(58,0.877)(59,0.876)(60,0.877)};\addlegendentry{Vanilla train}
       \addplot[blue,dashed]coordinates{(1,0.496)(2,0.667)(3,0.733)(4,0.765)(5,0.788)(6,0.807)(7,0.817)(8,0.827)(9,0.837)(10,0.843)(11,0.848)(12,0.850)(13,0.856)(14,0.859)(15,0.860)(16,0.865)(17,0.868)(18,0.870)(19,0.872)(20,0.871)(21,0.875)(22,0.876)(23,0.877)(24,0.878)(25,0.878)(26,0.879)(27,0.882)(28,0.882)(29,0.884)(30,0.884)(31,0.886)(32,0.884)(33,0.886)(34,0.890)(35,0.886)(36,0.889)(37,0.890)(38,0.889)(39,0.891)(40,0.892)(41,0.890)(42,0.892)(43,0.893)(44,0.892)(45,0.894)(46,0.895)(47,0.893)(48,0.895)(49,0.896)(50,0.895)(51,0.897)(52,0.894)(53,0.898)(54,0.896)(55,0.896)(56,0.896)(57,0.897)(58,0.896)(59,0.897)(60,0.898)};\addlegendentry{2/20 train}
         \addplot[red,dashed]coordinates{(1,0.445)(2,0.645)(3,0.733)(4,0.774)(5,0.796)(6,0.813)(7,0.825)(8,0.834)(9,0.843)(10,0.850)(11,0.853)(12,0.857)(13,0.862)(14,0.865)(15,0.868)(16,0.872)(17,0.875)(18,0.876)(19,0.879)(20,0.881)(21,0.882)(22,0.884)(23,0.886)(24,0.888)(25,0.889)(26,0.889)(27,0.891)(28,0.892)(29,0.893)(30,0.895)(31,0.894)(32,0.896)(33,0.896)(34,0.897)(35,0.898)(36,0.899)(37,0.898)(38,0.900)(39,0.898)(40,0.901)(41,0.900)(42,0.901)(43,0.902)(44,0.904)(45,0.904)(46,0.903)(47,0.903)(48,0.905)(49,0.905)(50,0.904)(51,0.906)(52,0.906)(53,0.908)(54,0.905)(55,0.907)(56,0.908)(57,0.909)(58,0.909)(59,0.909)(60,0.908)};\addlegendentry{5/500 train}
    \addplot[black]coordinates{(1,0.502)(2,0.617)(3,0.690)(4,0.687)(5,0.678)(6,0.749)(7,0.756)(8,0.729)(9,0.764)(10,0.743)(11,0.781)(12,0.811)(13,0.777)(14,0.800)(15,0.791)(16,0.690)(17,0.796)(18,0.762)(19,0.767)(20,0.779)(21,0.820)(22,0.782)(23,0.786)(24,0.815)(25,0.831)(26,0.794)(27,0.816)(28,0.815)(29,0.835)(30,0.784)(31,0.798)(32,0.806)(33,0.802)(34,0.810)(35,0.807)(36,0.737)(37,0.795)(38,0.836)(39,0.845)(40,0.795)(41,0.772)(42,0.813)(43,0.812)(44,0.804)(45,0.821)(46,0.827)(47,0.812)(48,0.772)(49,0.801)(50,0.823)(51,0.829)(52,0.813)(53,0.843)(54,0.837)(55,0.785)(56,0.831)(57,0.814)(58,0.827)(59,0.827)(60,0.840)};\addlegendentry{Vanilla test}
    \addplot[blue]coordinates{(1,0.579)(2,0.559)(3,0.688)(4,0.734)(5,0.749)(6,0.761)(7,0.779)(8,0.819)(9,0.797)(10,0.807)(11,0.797)(12,0.846)(13,0.819)(14,0.836)(15,0.832)(16,0.801)(17,0.849)(18,0.815)(19,0.813)(20,0.790)(21,0.853)(22,0.845)(23,0.841)(24,0.829)(25,0.839)(26,0.847)(27,0.857)(28,0.829)(29,0.832)(30,0.847)(31,0.860)(32,0.859)(33,0.852)(34,0.840)(35,0.849)(36,0.830)(37,0.864)(38,0.847)(39,0.867)(40,0.867)(41,0.872)(42,0.851)(43,0.862)(44,0.844)(45,0.870)(46,0.859)(47,0.849)(48,0.865)(49,0.867)(50,0.834)(51,0.857)(52,0.878)(53,0.849)(54,0.879)(55,0.855)(56,0.835)(57,0.855)(58,0.835)(59,0.855)(60,0.863)};\addlegendentry{2/20 test}
    \addplot[red]coordinates{(1,0.427)(2,0.601)(3,0.646)(4,0.730)(5,0.756)(6,0.788)(7,0.782)(8,0.805)(9,0.821)(10,0.804)(11,0.831)(12,0.802)(13,0.831)(14,0.846)(15,0.842)(16,0.827)(17,0.861)(18,0.838)(19,0.860)(20,0.851)(21,0.820)(22,0.858)(23,0.882)(24,0.860)(25,0.868)(26,0.864)(27,0.874)(28,0.868)(29,0.876)(30,0.850)(31,0.862)(32,0.872)(33,0.878)(34,0.860)(35,0.868)(36,0.874)(37,0.856)(38,0.881)(39,0.887)(40,0.889)(41,0.858)(42,0.884)(43,0.881)(44,0.876)(45,0.868)(46,0.877)(47,0.869)(48,0.867)(49,0.886)(50,0.873)(51,0.867)(52,0.884)(53,0.890)(54,0.892)(55,0.895)(56,0.877)(57,0.882)(58,0.886)(59,0.885)(60,0.890)};\addlegendentry{5/500 test}
    \end{axis}
    \end{tikzpicture}
    \caption{An extreme case of VNNE with 5 physical and 500 virtual networks, based on ResNet-20.}
    \label{fig-500}
\end{figure}
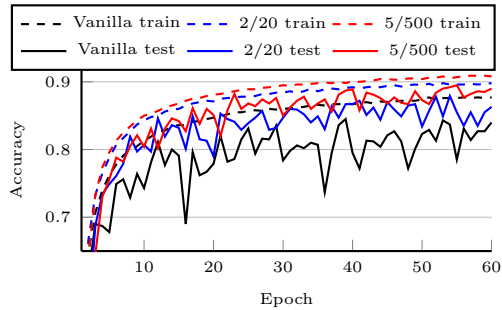

\textbf{Huge number of virtual neural networks.}
Figure~\ref{fig-500} visualizes an extreme case of training Resnet-20 VNNE that consists of 5 physical and 500 virtual neural networks with $f_A$ trained on CIFAR-10. The rest of the setting is identical to that in Section~\ref{sec-benchmark}. Training 500 networks simultaneously requires a significant amount of computational time. Therefore, the process was halted after 60 epochs and is presented here solely as a curiosity.
 However, it is obvious that the increase in physical and virtual capacity does not lead to overfitting but rather to a further increase in the accuracy of the test.


\section{Dead ends, limitations, and future work}
\label{sec-dead-ends}

\subsection{Equidistant generation of virtual networks}
Since the virtual networks in VNNE are generated randomly, they may, in extremely rare cases, be identical. To ensure maximal diversity in the inner structure, an algorithm was developed.  
Consider the sequence of $b$ blocks, denoted as $\sigma = \{s_1, s_2, \dots, s_b\}$, used for constructing a virtual neural network, where $s_i$ represents the physical network from which block $i$ is taken. Then, the generating sequence for the $j^{th}$ virtual network is determined as 
\begin{equation}
\mathbf{\sigma}_j = \mathbf{1}+t\left(j\frac{p^b}{v+1}\right), j \in \{1, 2, \dots, v\},
\end{equation}
where $t$ is a transformation function of the base-10 numeral system into the base-$p$ ($p$ is number of physical neural networks) numeral system with a length of $b$ characters. Such an algorithm produces the generating sequences where the sequences are spread equidistantly over the whole space of all possible combinations.
Example: having $p=2$, $b=6$, and $v=4$, the algorithm produces $\sigma_1 = 112211$, $\sigma_2 = 122112$, $\sigma_3 = 211221$, $\sigma_4 = 221122$, where it is obvious that at a certain position, all physical models are used exactly twice, not less, not more. In experiments, VNNE trained using this scheme reached worse results than a baseline and therefore, we abandoned it. Unfortunately, we did not find a proper explanation of why it failed.

\subsection{Diversity loss}
\label{subsec-diversity-loss}
Motivated by~\cite{zhang2020diversified}, we involved diversity loss in our compound loss, Eq.~\eqref{full-loss}, but the model was stacked with accuracy equal to random selection and could not be trained. The requirement of diversity prevents particular virtual networks from producing the correct prediction. In our case, all partial virtual networks are trained simultaneously, and diversity arises naturally (see Table~\ref{table-cifar-resnet}) when it is needed. In other words,instead of forcing partial networks (i.e., bring external bias) to yield high diversity and hope to high ensemble score, relying on inductive bias and focus on the ensemble score itself is more effective.

\subsection{Continuous replacement of non-promising inner networks}

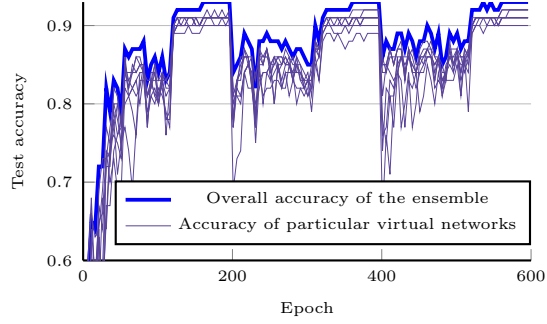
\begin{figure}
    \centering
    \begin{tikzpicture}
    \begin{axis}[height=50mm,width=75mm,legend columns=1, xlabel={Epoch},ylabel={Test accuracy},xmin=0, xmax=600, ymin=0.6, ymax=0.93,
    ymajorgrids=true,yminorgrids=true,style={thick},legend style={at={(1.0,0.31)}},axis x line*=bottom,axis y line*=left, style={font=\tiny}]
         \addplot[blue, line width=0.5mm]coordinates{(1,0.43)(6,0.56)(11,0.65)(16,0.64)(21,0.72)(26,0.72)(31,0.82)(36,0.79)(41,0.83)(46,0.82)(51,0.79)(56,0.87)(61,0.86)(66,0.87)(71,0.87)(76,0.87)(81,0.88)(86,0.83)(91,0.85)(96,0.86)(101,0.84)(106,0.86)(111,0.83)(116,0.84)(121,0.91)(126,0.92)(131,0.92)(136,0.92)(141,0.92)(146,0.92)(151,0.92)(156,0.92)(161,0.93)(166,0.93)(171,0.93)(176,0.93)(181,0.93)(186,0.93)(191,0.93)(196,0.93)(201,0.84)(206,0.85)(211,0.86)(216,0.89)(221,0.88)(226,0.87)(231,0.82)(236,0.88)(241,0.88)(246,0.89)(251,0.88)(256,0.88)(261,0.88)(266,0.87)(271,0.88)(276,0.87)(281,0.86)(286,0.87)(291,0.87)(296,0.86)(301,0.85)(306,0.85)(311,0.90)(316,0.88)(321,0.91)(326,0.92)(331,0.92)(336,0.92)(341,0.92)(346,0.92)(351,0.92)(356,0.92)(361,0.93)(366,0.93)(371,0.93)(376,0.93)(381,0.93)(386,0.93)(391,0.93)(396,0.93)(401,0.86)(406,0.88)(411,0.87)(416,0.87)(421,0.89)(426,0.89)(431,0.87)(436,0.89)(441,0.88)(446,0.86)(451,0.88)(456,0.88)(461,0.90)(466,0.89)(471,0.87)(476,0.89)(481,0.85)(486,0.86)(491,0.89)(496,0.88)(501,0.89)(506,0.87)(511,0.88)(516,0.88)(521,0.92)(526,0.92)(531,0.92)(536,0.93)(541,0.92)(546,0.93)(551,0.92)(556,0.92)(561,0.93)(566,0.93)(571,0.93)(576,0.93)(581,0.93)(586,0.93)(591,0.93)(596,0.93)};\addlegendentry{Overall accuracy of the ensemble}
    \addplot[Violet, line width=0.1mm]coordinates{(1,0.34)(6,0.58)(11,0.68)(16,0.56)(21,0.67)(26,0.64)(31,0.72)(36,0.72)(41,0.80)(46,0.80)(51,0.74)(56,0.86)(61,0.82)(66,0.85)(71,0.84)(76,0.85)(81,0.85)(86,0.81)(91,0.83)(96,0.83)(101,0.83)(106,0.83)(111,0.81)(116,0.82)(121,0.90)(126,0.90)(131,0.91)(136,0.91)(141,0.90)(146,0.90)(151,0.90)(156,0.90)(161,0.91)(166,0.91)(171,0.91)(176,0.91)(181,0.91)(186,0.91)(191,0.91)(196,0.91)(201,0.80)(206,0.82)(211,0.86)(216,0.86)(221,0.86)(226,0.83)(231,0.78)(236,0.86)(241,0.86)(246,0.86)(251,0.85)(256,0.86)(261,0.84)(266,0.85)(271,0.86)(276,0.85)(281,0.83)(286,0.86)(291,0.83)(296,0.83)(301,0.81)(306,0.83)(311,0.88)(316,0.86)(321,0.90)(326,0.91)(331,0.90)(336,0.90)(341,0.91)(346,0.90)(351,0.90)(356,0.90)(361,0.91)(366,0.91)(371,0.91)(376,0.91)(381,0.91)(386,0.91)(391,0.91)(396,0.91)(401,0.85)(406,0.85)(411,0.87)(416,0.85)(421,0.87)(426,0.87)(431,0.83)(436,0.87)(441,0.86)(446,0.84)(451,0.87)(456,0.88)(461,0.88)(466,0.87)(471,0.84)(476,0.86)(481,0.83)(486,0.83)(491,0.87)(496,0.87)(501,0.87)(506,0.84)(511,0.87)(516,0.86)(521,0.90)(526,0.91)(531,0.91)(536,0.91)(541,0.91)(546,0.91)(551,0.91)(556,0.90)(561,0.91)(566,0.91)(571,0.91)(576,0.91)(581,0.91)(586,0.91)(591,0.91)(596,0.91)};\addlegendentry{Accuracy of particular virtual networks}
    \addplot[Violet, line width=0.1mm]coordinates{(1,0.29)(6,0.39)(11,0.64)(16,0.58)(21,0.65)(26,0.66)(31,0.74)(36,0.73)(41,0.77)(46,0.80)(51,0.78)(56,0.84)(61,0.82)(66,0.84)(71,0.84)(76,0.85)(81,0.85)(86,0.81)(91,0.82)(96,0.84)(101,0.80)(106,0.83)(111,0.83)(116,0.82)(121,0.89)(126,0.90)(131,0.90)(136,0.90)(141,0.90)(146,0.90)(151,0.90)(156,0.90)(161,0.91)(166,0.91)(171,0.91)(176,0.90)(181,0.90)(186,0.91)(191,0.90)(196,0.91)(201,0.78)(206,0.83)(211,0.85)(216,0.86)(221,0.86)(226,0.82)(231,0.77)(236,0.85)(241,0.85)(246,0.87)(251,0.85)(256,0.86)(261,0.85)(266,0.85)(271,0.87)(276,0.86)(281,0.83)(286,0.85)(291,0.84)(296,0.82)(301,0.81)(306,0.82)(311,0.88)(316,0.87)(321,0.89)(326,0.90)(331,0.90)(336,0.90)(341,0.90)(346,0.90)(351,0.90)(356,0.90)(361,0.91)(366,0.91)(371,0.90)(376,0.91)(381,0.91)(386,0.91)(391,0.91)(396,0.91)(401,0.85)(406,0.85)(411,0.87)(416,0.85)(421,0.87)(426,0.87)(431,0.83)(436,0.87)(441,0.84)(446,0.84)(451,0.86)(456,0.87)(461,0.87)(466,0.86)(471,0.85)(476,0.86)(481,0.83)(486,0.84)(491,0.86)(496,0.86)(501,0.86)(506,0.84)(511,0.86)(516,0.86)(521,0.90)(526,0.90)(531,0.90)(536,0.91)(541,0.91)(546,0.91)(551,0.90)(556,0.90)(561,0.91)(566,0.91)(571,0.91)(576,0.91)(581,0.91)(586,0.91)(591,0.91)(596,0.91)};
    \addplot[Violet, line width=0.1mm]coordinates{(1,0.31)(6,0.44)(11,0.32)(16,0.47)(21,0.63)(26,0.53)(31,0.67)(36,0.67)(41,0.74)(46,0.76)(51,0.78)(56,0.80)(61,0.81)(66,0.79)(71,0.76)(76,0.83)(81,0.83)(86,0.78)(91,0.81)(96,0.80)(101,0.82)(106,0.82)(111,0.78)(116,0.78)(121,0.89)(126,0.89)(131,0.89)(136,0.89)(141,0.90)(146,0.89)(151,0.89)(156,0.89)(161,0.90)(166,0.90)(171,0.90)(176,0.90)(181,0.90)(186,0.90)(191,0.90)(196,0.90)(201,0.69)(206,0.73)(211,0.74)(216,0.84)(221,0.82)(226,0.82)(231,0.84)(236,0.85)(241,0.80)(246,0.84)(251,0.85)(256,0.80)(261,0.85)(266,0.83)(271,0.85)(276,0.85)(281,0.84)(286,0.80)(291,0.82)(296,0.81)(301,0.82)(306,0.85)(311,0.85)(316,0.83)(321,0.88)(326,0.89)(331,0.89)(336,0.88)(341,0.89)(346,0.88)(351,0.87)(356,0.88)(361,0.88)(366,0.89)(371,0.89)(376,0.88)(381,0.89)(386,0.89)(391,0.89)(396,0.89)(401,0.66)(406,0.76)(411,0.71)(416,0.81)(421,0.86)(426,0.77)(431,0.80)(436,0.84)(441,0.77)(446,0.80)(451,0.86)(456,0.85)(461,0.87)(466,0.88)(471,0.86)(476,0.86)(481,0.85)(486,0.83)(491,0.88)(496,0.85)(501,0.87)(506,0.81)(511,0.87)(516,0.86)(521,0.91)(526,0.91)(531,0.91)(536,0.91)(541,0.91)(546,0.91)(551,0.91)(556,0.91)(561,0.91)(566,0.91)(571,0.92)(576,0.92)(581,0.92)(586,0.92)(591,0.92)(596,0.92)};
    \addplot[Violet, line width=0.1mm]coordinates{(1,0.24)(6,0.38)(11,0.52)(16,0.56)(21,0.64)(26,0.53)(31,0.74)(36,0.70)(41,0.73)(46,0.69)(51,0.69)(56,0.82)(61,0.81)(66,0.82)(71,0.81)(76,0.82)(81,0.85)(86,0.77)(91,0.83)(96,0.81)(101,0.80)(106,0.83)(111,0.78)(116,0.79)(121,0.90)(126,0.91)(131,0.90)(136,0.91)(141,0.91)(146,0.91)(151,0.91)(156,0.90)(161,0.91)(166,0.91)(171,0.91)(176,0.91)(181,0.91)(186,0.91)(191,0.91)(196,0.91)(201,0.81)(206,0.81)(211,0.80)(216,0.84)(221,0.85)(226,0.84)(231,0.75)(236,0.83)(241,0.84)(246,0.84)(251,0.83)(256,0.84)(261,0.85)(266,0.82)(271,0.84)(276,0.84)(281,0.84)(286,0.84)(291,0.84)(296,0.83)(301,0.84)(306,0.81)(311,0.87)(316,0.86)(321,0.90)(326,0.90)(331,0.91)(336,0.90)(341,0.90)(346,0.91)(351,0.91)(356,0.90)(361,0.91)(366,0.91)(371,0.91)(376,0.91)(381,0.91)(386,0.91)(391,0.91)(396,0.91)(401,0.74)(406,0.82)(411,0.81)(416,0.81)(421,0.84)(426,0.86)(431,0.84)(436,0.86)(441,0.87)(446,0.85)(451,0.85)(456,0.86)(461,0.86)(466,0.86)(471,0.83)(476,0.86)(481,0.82)(486,0.84)(491,0.87)(496,0.85)(501,0.86)(506,0.85)(511,0.85)(516,0.84)(521,0.90)(526,0.91)(531,0.91)(536,0.91)(541,0.91)(546,0.91)(551,0.91)(556,0.91)(561,0.92)(566,0.92)(571,0.92)(576,0.92)(581,0.92)(586,0.92)(591,0.92)(596,0.92)};
    \addplot[Violet, line width=0.1mm]coordinates{(1,0.34)(6,0.47)(11,0.51)(16,0.47)(21,0.69)(26,0.73)(31,0.73)(36,0.77)(41,0.80)(46,0.78)(51,0.75)(56,0.85)(61,0.83)(66,0.84)(71,0.86)(76,0.86)(81,0.84)(86,0.81)(91,0.82)(96,0.85)(101,0.83)(106,0.84)(111,0.82)(116,0.82)(121,0.90)(126,0.91)(131,0.91)(136,0.91)(141,0.91)(146,0.91)(151,0.91)(156,0.91)(161,0.91)(166,0.92)(171,0.91)(176,0.92)(181,0.91)(186,0.92)(191,0.91)(196,0.92)(201,0.81)(206,0.83)(211,0.84)(216,0.87)(221,0.86)(226,0.86)(231,0.77)(236,0.87)(241,0.85)(246,0.87)(251,0.83)(256,0.86)(261,0.86)(266,0.86)(271,0.85)(276,0.86)(281,0.85)(286,0.85)(291,0.85)(296,0.84)(301,0.83)(306,0.83)(311,0.88)(316,0.87)(321,0.90)(326,0.91)(331,0.91)(336,0.91)(341,0.91)(346,0.91)(351,0.91)(356,0.91)(361,0.92)(366,0.92)(371,0.92)(376,0.92)(381,0.92)(386,0.92)(391,0.92)(396,0.92)(401,0.76)(406,0.77)(411,0.83)(416,0.82)(421,0.85)(426,0.86)(431,0.85)(436,0.86)(441,0.85)(446,0.84)(451,0.84)(456,0.85)(461,0.86)(466,0.86)(471,0.85)(476,0.86)(481,0.84)(486,0.84)(491,0.87)(496,0.84)(501,0.87)(506,0.83)(511,0.85)(516,0.87)(521,0.90)(526,0.91)(531,0.91)(536,0.91)(541,0.91)(546,0.91)(551,0.91)(556,0.91)(561,0.92)(566,0.92)(571,0.92)(576,0.92)(581,0.92)(586,0.92)(591,0.92)(596,0.92)};
    \addplot[Violet, line width=0.1mm]coordinates{(1,0.38)(6,0.50)(11,0.63)(16,0.47)(21,0.55)(26,0.68)(31,0.64)(36,0.74)(41,0.78)(46,0.71)(51,0.75)(56,0.77)(61,0.74)(66,0.69)(71,0.77)(76,0.84)(81,0.84)(86,0.77)(91,0.84)(96,0.84)(101,0.80)(106,0.83)(111,0.83)(116,0.84)(121,0.89)(126,0.90)(131,0.90)(136,0.90)(141,0.91)(146,0.91)(151,0.90)(156,0.90)(161,0.91)(166,0.91)(171,0.91)(176,0.91)(181,0.91)(186,0.91)(191,0.91)(196,0.91)(201,0.77)(206,0.83)(211,0.83)(216,0.87)(221,0.85)(226,0.86)(231,0.83)(236,0.87)(241,0.86)(246,0.87)(251,0.84)(256,0.85)(261,0.86)(266,0.86)(271,0.85)(276,0.85)(281,0.86)(286,0.84)(291,0.84)(296,0.83)(301,0.83)(306,0.83)(311,0.87)(316,0.87)(321,0.90)(326,0.90)(331,0.91)(336,0.91)(341,0.91)(346,0.90)(351,0.91)(356,0.90)(361,0.91)(366,0.91)(371,0.91)(376,0.91)(381,0.91)(386,0.91)(391,0.91)(396,0.91)(401,0.84)(406,0.84)(411,0.86)(416,0.86)(421,0.86)(426,0.86)(431,0.84)(436,0.86)(441,0.84)(446,0.82)(451,0.84)(456,0.84)(461,0.86)(466,0.86)(471,0.82)(476,0.83)(481,0.82)(486,0.85)(491,0.86)(496,0.85)(501,0.87)(506,0.86)(511,0.84)(516,0.86)(521,0.90)(526,0.91)(531,0.91)(536,0.91)(541,0.91)(546,0.91)(551,0.90)(556,0.91)(561,0.91)(566,0.91)(571,0.91)(576,0.91)(581,0.91)(586,0.91)(591,0.91)(596,0.91)};
    \addplot[Violet, line width=0.1mm]coordinates{(1,0.30)(6,0.51)(11,0.35)(16,0.51)(21,0.63)(26,0.56)(31,0.65)(36,0.75)(41,0.79)(46,0.76)(51,0.77)(56,0.85)(61,0.81)(66,0.82)(71,0.83)(76,0.84)(81,0.84)(86,0.81)(91,0.83)(96,0.81)(101,0.82)(106,0.81)(111,0.79)(116,0.77)(121,0.90)(126,0.91)(131,0.90)(136,0.90)(141,0.91)(146,0.91)(151,0.90)(156,0.90)(161,0.91)(166,0.91)(171,0.91)(176,0.91)(181,0.91)(186,0.91)(191,0.91)(196,0.91)(201,0.82)(206,0.81)(211,0.81)(216,0.84)(221,0.85)(226,0.84)(231,0.80)(236,0.82)(241,0.84)(246,0.84)(251,0.85)(256,0.80)(261,0.85)(266,0.82)(271,0.82)(276,0.85)(281,0.83)(286,0.82)(291,0.82)(296,0.82)(301,0.83)(306,0.82)(311,0.86)(316,0.83)(321,0.89)(326,0.90)(331,0.90)(336,0.90)(341,0.90)(346,0.90)(351,0.90)(356,0.90)(361,0.90)(366,0.91)(371,0.90)(376,0.91)(381,0.91)(386,0.91)(391,0.91)(396,0.91)(401,0.78)(406,0.75)(411,0.81)(416,0.77)(421,0.84)(426,0.83)(431,0.77)(436,0.83)(441,0.82)(446,0.84)(451,0.79)(456,0.81)(461,0.85)(466,0.83)(471,0.80)(476,0.84)(481,0.81)(486,0.83)(491,0.83)(496,0.83)(501,0.85)(506,0.81)(511,0.86)(516,0.79)(521,0.89)(526,0.90)(531,0.90)(536,0.89)(541,0.90)(546,0.90)(551,0.90)(556,0.90)(561,0.90)(566,0.90)(571,0.90)(576,0.90)(581,0.90)(586,0.90)(591,0.90)(596,0.90)};
    \addplot[Violet, line width=0.1mm]coordinates{(1,0.41)(6,0.60)(11,0.66)(16,0.63)(21,0.64)(26,0.69)(31,0.77)(36,0.75)(41,0.76)(46,0.75)(51,0.73)(56,0.83)(61,0.83)(66,0.82)(71,0.83)(76,0.80)(81,0.84)(86,0.79)(91,0.81)(96,0.82)(101,0.80)(106,0.84)(111,0.76)(116,0.82)(121,0.89)(126,0.90)(131,0.90)(136,0.90)(141,0.90)(146,0.90)(151,0.91)(156,0.90)(161,0.90)(166,0.91)(171,0.91)(176,0.91)(181,0.91)(186,0.91)(191,0.91)(196,0.91)(201,0.77)(206,0.82)(211,0.81)(216,0.84)(221,0.85)(226,0.83)(231,0.79)(236,0.84)(241,0.81)(246,0.82)(251,0.82)(256,0.83)(261,0.83)(266,0.82)(271,0.84)(276,0.81)(281,0.83)(286,0.85)(291,0.83)(296,0.82)(301,0.83)(306,0.79)(311,0.85)(316,0.86)(321,0.90)(326,0.90)(331,0.90)(336,0.89)(341,0.89)(346,0.89)(351,0.90)(356,0.89)(361,0.90)(366,0.90)(371,0.90)(376,0.90)(381,0.90)(386,0.90)(391,0.90)(396,0.90)(401,0.81)(406,0.85)(411,0.84)(416,0.83)(421,0.84)(426,0.86)(431,0.86)(436,0.86)(441,0.85)(446,0.84)(451,0.86)(456,0.86)(461,0.87)(466,0.86)(471,0.83)(476,0.86)(481,0.81)(486,0.84)(491,0.85)(496,0.84)(501,0.86)(506,0.86)(511,0.86)(516,0.85)(521,0.89)(526,0.90)(531,0.91)(536,0.91)(541,0.90)(546,0.91)(551,0.91)(556,0.90)(561,0.91)(566,0.91)(571,0.91)(576,0.91)(581,0.91)(586,0.91)(591,0.91)(596,0.91)};
    \addplot[Violet, line width=0.1mm]coordinates{(1,0.26)(6,0.41)(11,0.59)(16,0.60)(21,0.59)(26,0.59)(31,0.73)(36,0.74)(41,0.82)(46,0.81)(51,0.77)(56,0.84)(61,0.84)(66,0.84)(71,0.86)(76,0.85)(81,0.84)(86,0.83)(91,0.83)(96,0.85)(101,0.81)(106,0.84)(111,0.83)(116,0.84)(121,0.90)(126,0.91)(131,0.91)(136,0.91)(141,0.91)(146,0.91)(151,0.90)(156,0.90)(161,0.91)(166,0.91)(171,0.91)(176,0.91)(181,0.91)(186,0.91)(191,0.91)(196,0.91)(201,0.78)(206,0.83)(211,0.84)(216,0.86)(221,0.86)(226,0.82)(231,0.78)(236,0.86)(241,0.85)(246,0.87)(251,0.85)(256,0.85)(261,0.86)(266,0.85)(271,0.86)(276,0.85)(281,0.84)(286,0.86)(291,0.85)(296,0.82)(301,0.83)(306,0.84)(311,0.87)(316,0.87)(321,0.90)(326,0.91)(331,0.91)(336,0.91)(341,0.91)(346,0.91)(351,0.91)(356,0.91)(361,0.91)(366,0.91)(371,0.91)(376,0.92)(381,0.92)(386,0.92)(391,0.92)(396,0.92)(401,0.86)(406,0.84)(411,0.87)(416,0.85)(421,0.87)(426,0.88)(431,0.84)(436,0.87)(441,0.84)(446,0.84)(451,0.86)(456,0.86)(461,0.87)(466,0.87)(471,0.86)(476,0.85)(481,0.83)(486,0.84)(491,0.86)(496,0.86)(501,0.87)(506,0.86)(511,0.86)(516,0.87)(521,0.90)(526,0.91)(531,0.91)(536,0.91)(541,0.91)(546,0.91)(551,0.91)(556,0.91)(561,0.91)(566,0.91)(571,0.91)(576,0.91)(581,0.91)(586,0.91)(591,0.91)(596,0.91)};
    \end{axis}
    \end{tikzpicture}
    \caption{Impact of replacing two worst networks in the VNNE. Two worst network in the VNNE are replaced each 200 epochs followed by learning rate restart and VNNE recompilation.
    }
    \label{fig-replacing-not-promising}
\end{figure}

The performance of individual virtual neural networks within the ensemble was analyzed, revealing that their accuracies were not consistent. Consequently, training was conducted with a restarted learning rate, followed by VNNE recompilation every 200 epochs. Prior to each recompilation, the two virtual networks with the highest training loss were replaced.
 For training, the same settings as described in Section~\ref{sec-benchmark} were used. Figure~\ref{fig-replacing-not-promising} shows an illustration of such a process with two restarts of the learning rate and the removal of two worst subnetworks. Since the proposed scheme did not enhance the overall performance of VNNE, it is assumed that even subnetworks with lower performance contribute to the ensemble and remain useful.

\subsection{Application to Transformer-based architectures}
\label{subsec-transformers}
The proposed virtualization scheme requires a split architecture into blocks and their fusion strategy. The first point is fulfilled by many transformer-based architectures as Swin, Eva, MaxVit etc. Regarding the second requirement, there is no theoretical justification that adding/multiplication of output from various transformer blocks will work, and deeper research, which is beyond the focus of our paper, is necessary.

\section{Conclusion}
A new paradigm has been introduced that scales computational cost while maintaining a fixed model capacity. This paradigm is based on a novel ensembling approach, where a large number of tangled siamese networks share the weights of a few physical models. The tangled structure facilitates an intrinsic form of combination and distortion, leading to higher accuracy compared to both the baseline model and the deep ensemble principle.  
An important property of VNNE is that it functions not only as an ensemble but also as a method for improving model training. Experimental results demonstrate that the best single model from VNNE outperforms models trained using conventional methods and even surpasses models with larger capacity. This effect is particularly pronounced in small models, whereas medium to large models show a lower or negligible improvement. While the performance boost may be relatively small, the key observation is that the best single model from VNNE consistently achieves higher accuracy than the best model trained using standard techniques. This holds even when the VNNE-trained model is compared to standard versions with greater capacity.

\section*{Acknowledgment}
The study described is from the project "Research of Excellence on Digital Technologies and Wellbeing CZ.02.01.01/00/22\_008/0004583" which is co-financed by the European Union.

%
\section*{Conflict of interest}
The authors declare that they have no conflict of interest.

\section*{Compliance with Ethical Standards}
This article does not contain any studies with human participants or animals performed by any of the authors.

\bibliographystyle{unsrt}
\bibliography{peters_db}

\end{document}